\documentclass{article} 
\usepackage{iclr2026_conference,times}

\usepackage{amsmath,amsfonts,bm}

\def\eqref#1{equation~\ref{#1}}

\def\1{\bm{1}}

\DeclareMathAlphabet{\mathsfit}{\encodingdefault}{\sfdefault}{m}{sl}
\SetMathAlphabet{\mathsfit}{bold}{\encodingdefault}{\sfdefault}{bx}{n}

\usepackage{hyperref}
\usepackage{url}
\usepackage{graphicx} 
\usepackage{subcaption}
\usepackage{url}
\usepackage[utf8]{inputenc} 
\usepackage[T1]{fontenc}    
\usepackage{booktabs}       
\usepackage{amsfonts}       
\usepackage{nicefrac}       
\usepackage{microtype}      
\usepackage{xcolor}         
\usepackage{graphicx}
\usepackage{wrapfig}
\usepackage{multirow}
\usepackage{epsfig}
\usepackage{amsmath}
\usepackage{amssymb}
\usepackage{marvosym}
\usepackage{soul,color}
\usepackage{enumitem}
\usepackage{threeparttable}
\usepackage{algorithm}
\usepackage{algpseudocode}
\usepackage{setspace}
\usepackage{amsthm}
\usepackage{dutchcal}
\usepackage{makecell}
\usepackage{pifont}
\usepackage{xspace}
\usepackage{colortbl}

\usepackage{algorithm}
\usepackage{algpseudocode}

\usepackage{tocloft}

\usepackage{diagbox}        

\usepackage{multirow}
\usepackage{float}

\usepackage{makecell}

\usepackage{amsmath}
\usepackage{etoc}

\newcommand{\cut}[1]{}
\newcommand{\std}[1]{\scriptsize{$\pm$#1}}

\title{Channel-Imposed Fusion: A Simple yet Effective
Method for Medical Time Series Classification}

\author{
    \textbf{Ming Hu}$^{1,2}$ \hspace{1em}
    \textbf{Jianfu Yin}$^{1,2}$ \hspace{1em}
    \textbf{Mingyu Dou}$^{1,2}$ 
    \textbf{Yuqi Wang}$^{1,2}$ \hspace{1em}
    \textbf{Ruochen Dang}$^{1,2}$ \hspace{1em} \\
    \textbf{Siyi Liang}$^{3}$  \hspace{1em}
    \textbf{Feiyu Zhu}$^{6}$  \hspace{1em}
    \textbf{Cong Hu}$^{4}$ \hspace{1em}
    \textbf{Yao Wang}$^{5}$\thanks{Corresponding author.} \hspace{1em}
    \textbf{Bingliang Hu}$^1$\footnotemark[1] \hspace{1em}
    \textbf{Quan Wang}$^1$\footnotemark[1] \hspace{1em} \\
    \small
    $^{1}$Xi'an Institute of Optics and Precision Mechanics, CAS \\
    $^{2}$University of Chinese Academy of Sciences
    $^{3}$Southwest Jiaotong University \\
    $^{4}$Zhongnan Hospital of Wuhan University
    $^{5}$Xi'an Jiaotong University \\
    $^{6}$National Key Laboratory of Information Systems Engineering \\
    huming708@gmail.com}

\iclrfinalcopy 
\begin{document}

\maketitle

\begin{abstract}

Medical time series (MedTS) such as EEG and ECG are critical for clinical diagnosis, yet existing deep learning approaches often struggle with two key challenges: the misalignment between domain-specific physiological knowledge and generic architectures, and the inherent low signal-to-noise ratio (SNR) of MedTS. To address these limitations, we shift from a conventional model-centric paradigm toward a data-centric perspective grounded in physiological principles. We propose Channel-Imposed Fusion (CIF), a method that explicitly encodes causal inter-channel relationships by linearly combining signals under domain-informed constraints, thereby enabling interpretable signal enhancement and noise suppression. To further demonstrate the effectiveness of data-centric design, we develop a simple yet powerful model, Hidden-layer Mixed Bidirectional Temporal Convolutional Network (HM-BiTCN), which, when combined with CIF, consistently outperforms Transformer-based approaches on multiple MedTS benchmarks and achieves new state-of-the-art performance on general time series classification datasets. Moreover, CIF is architecture-agnostic and can be seamlessly integrated into mainstream models such as Transformers, enhancing their adaptability to medical scenarios. Our work highlights the necessity of rethinking MedTS classification from a data-centric perspective and establishes a transferable framework for bridging physiological priors with modern deep learning architectures.
The complete source code supporting this study is publicly available at the following Link: 
\href{https://github.com/Xi-Mu-Yu/CIF}{https://github.com/Xi-Mu-Yu/CIF}.

\end{abstract}

\section{Introduction}

Medical time series (MedTS) data, such as electroencephalogram (EEG) and electrocardiogram (ECG) signals, are widely used in clinical settings to monitor patient health and play a crucial role in diagnosing neurological and cardiovascular diseases~\cite{arif2024ef, xiao2023deep, zhu2025mtnet, wang2024adformer, wang2025lead}. Accurate classification of these signals enables early anomaly detection, personalized treatment, and optimized therapy planning, ultimately improving patient outcomes and healthcare efficiency~\cite{liu2024classification, tian2023automatic}.
With advances in deep learning, CNN-based models like EEGNet \cite{lawhern2018eegnet} can automatically extract informative features from raw signals, significantly improving classification performance.

\begin{figure}[htt] 
    \centering
    \includegraphics[width=1\textwidth]{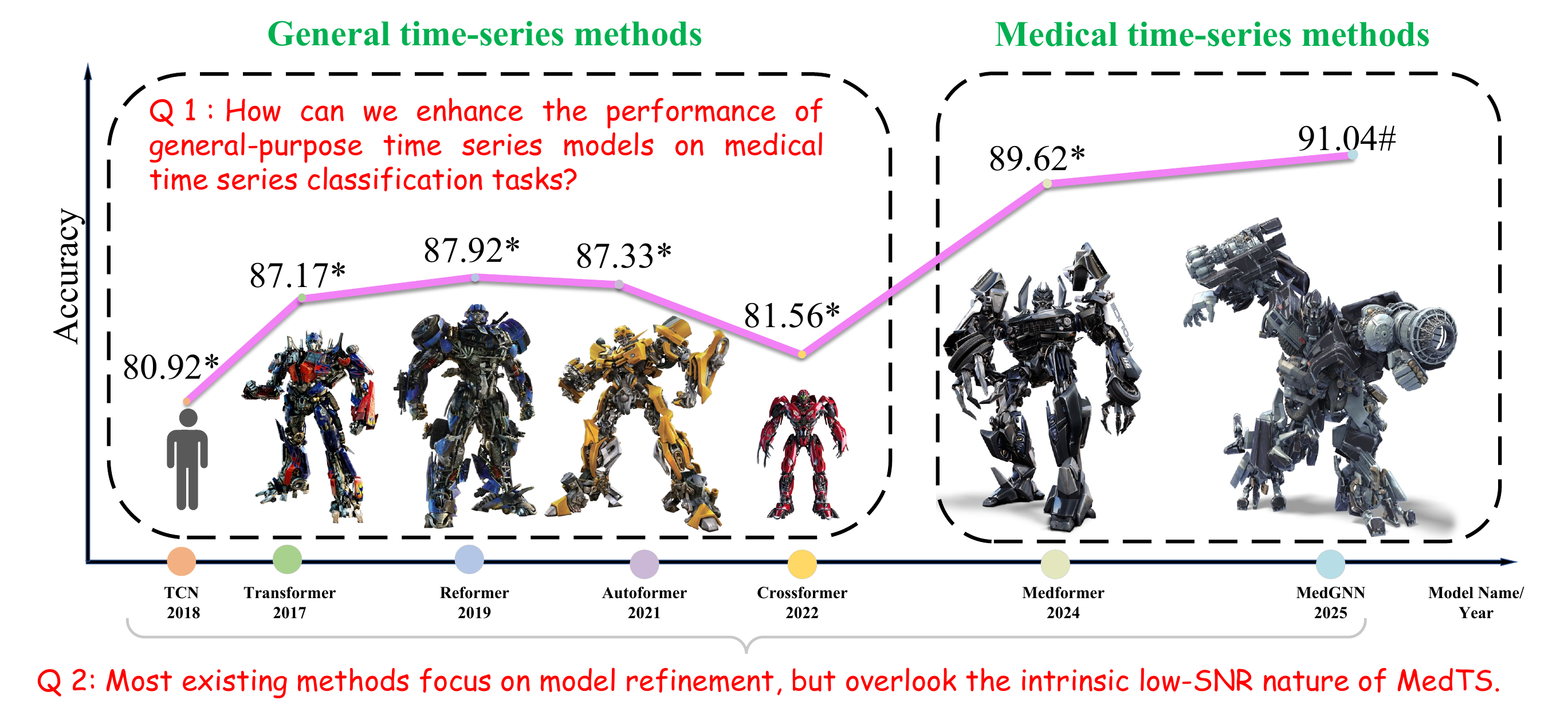} 
    \caption{The results of various methods on the TDBrain dataset (EEG) are presented, where
\textasteriskcentered~indicates results reported by \textit{Medformer~\cite{wang2024medformer}}, and 
\# indicates results reported by \textit{MedGNN~\cite{fan2025medgnn}}. In addition, we highlight 
two main motivations of this work (Q1 and Q2).}
    \label{fig:1}
\end{figure}

In recent years, Transformer models~\cite{vaswani2017attention}, originally inspired by the self-attention mechanism~\cite{bahdanau2014neural}, have achieved remarkable progress in time series modeling, particularly in capturing long-range dependencies and global contextual information~\cite{liu2021pyraformer, zhou2021informer}. By mapping sequential data into high-dimensional token embeddings, Transformers are able to implicitly model complex temporal dependencies. Despite their success across a wide range of time series tasks, applying Transformer architectures to MedTS classification still faces several challenges, which can be summarized as follows:
\textbf{(1) Misalignment between domain-specific knowledge and generic architectures.}  
Mainstream time series models, such as Autoformer~\cite{wu2021autoformer}, Crossformer~\cite{zhang2022crossformer}, and Reformer~\cite{kitaev2019reformer}, have demonstrated strong performance in general domains such as weather forecasting and finance. However, as illustrated in Fig.~\ref{fig:1}, these approaches fail to achieve comparable effectiveness in MedTS classification tasks. This raises the urgent question of how to enhance the applicability of general-purpose models in medical scenarios. Moreover, MedTS often encode critical physiological characteristics---for example, conduction delays across ECG leads~\cite{auricchio2014does} and rhythmic synchrony in EEG signals~\cite{palva2014correlation, fries2015rhythms}---which inherently reflect \emph{channel-level relationships}. Unfortunately, such physiological dependencies are rarely considered in generic time series modeling frameworks.
\textbf{(2) Overemphasis on model optimization while neglecting the intrinsic low SNR of MedTS.}  
Unlike general-purpose time series tasks, MedTS are characterized by pronounced low-SNR conditions~\cite{del2011assessment, sraitih2022robustness, sharma2017single, mohd2020analysis, jia2024preprocessing}, where noise and artifacts can easily overshadow critical physiological features. In such conditions, complex Transformer architectures do not always succeed in stably extracting effective representations, while simpler models (e.g., TCNs~\cite{bai2018empirical}) may also experience more severe performance degradation. Indeed, recent Transformer-based methods tailored for MedTS, such as MedGNN~\cite{fan2025medgnn} and Medformer~\cite{wang2024medformer}, primarily rely on architectural innovations, yet they fall short in fundamentally addressing the low-SNR challenge. This raises a key question: should breakthroughs in MedTS classification come from increasingly complex architectures, or from more principled data processing and representation strategies?

To address the aforementioned limitations, we depart from the traditional \emph{model-centric} paradigm that relies on increasingly complex architectures to capture temporal dependencies, and instead propose a \textbf{data-centric} approach grounded in the physiological properties of medical time series.  
Following this principle, we introduce the \textit{Channel-Imposed Fusion (CIF)} method, which explicitly encodes prior causal structures into feature representations. Specifically, CIF constructs new features through a linear combination of signals from different channels:
\[
x_{\text{new}} = a x + b y,
\]
where $x$ and $y$ denote signals from two distinct channels, and $a$ and $b$ are coefficients predefined based on domain knowledge.  
When $a$ and $b$ take fixed values, they are not learned directly from patient data, but instead derived from two domain-specific prior hypotheses:  
\textbf{(1) Physiological Coupling Hypothesis.}  
For ECG signals, when two leads are highly correlated (e.g., P-wave polarity and morphology are consistent~\cite{platonov2012p}), setting $a = b = 1$ achieves in-phase summation, thereby enhancing target signal components and improving the SNR.  
\textbf{(2) Noise Suppression Hypothesis.}  
In EEG recordings, ocular artifacts such as blinks often appear highly correlated in frontal electrodes Fp1 and Fp2~\cite{croft2000removal}. To suppress such noise, we set $a = 1, b = -1$, applying a differential fusion strategy to cancel common-mode interference.  
Here, the coefficients $a$ and $b$ serve as symbolic encodings of interpretable physiological principles, rather than exact data-driven estimates. When treated as learnable parameters, they can be fine-tuned under symbolic constraints imposed by prior knowledge (e.g., enforcing $a > 0, b > 0$ under coupling, and $a > 0, b < 0$ under noise suppression). This design maintains the interpretability of directional relationships (e.g., signal enhancement or cancellation) while allowing the model to adaptively adjust the magnitude of each coefficient based on the training data.

\textbf{To emphasize the importance of data-centric approaches}, we deliberately designed a simple yet effective model—the Hidden-layer Mixed Bidirectional Temporal Convolutional Network (HM-BiTCN)—to demonstrate that excellent performance does not necessarily require model complexity. The combination of CIF and HM-BiTCN not only outperforms Transformer-based methods on multiple medical datasets but also achieves new state-of-the-art (SOTA) results on general time series classification benchmarks. More importantly, the CIF method is not limited to the HM-BiTCN architecture itself; it exhibits strong transferability and can be seamlessly integrated into existing Transformer architectures, enhancing their adaptability to MedTS data.
Our main contributions are:

\begin{itemize}
    \item \textbf{Proposal of Channel-Imposed Fusion (CIF).}  
    We introduce CIF to explicitly model inter-channel relationships in medical time series, particularly suitable for signals with well-defined physiological structures such as EEG and ECG.  

    \item \textbf{Design of HM-BiTCN based on CIF.}  
    By integrating CIF into HM-BiTCN, our method consistently outperforms existing SOTA models across multiple publicly available medical and non-medical time series classification datasets.  

    \item \textbf{Methodological transferability.}  
    CIF is architecture-agnostic and can be seamlessly integrated into mainstream models such as Transformers, compensating for the limitations of traditional positional encodings in modeling channel-level correlations, and highlighting the paradigm shift from a \emph{model-centric} to a \emph{data-centric} perspective.

\end{itemize}

\section{Related Work}
\textbf{Medical Time Series Classification.}
Medical time series analysis diverges fundamentally from general time series forecasting \cite{wu2022timesnet,lu2024cats} by prioritizing pathological signature decoding over temporal extrapolation, with modalities like EEG \cite{tangself,yangmanydg,quensemble}, ECG \cite{xiao2023deep,wang2023hierarchical,kiyasseh2021clocs}, and EMG [\cite{xiong2021deep,dai2022mseva}] encoding distinct clinical semantics. Early methods were dominated by compact CNNs such as EEGNet~\cite{lawhern2018eegnet}, which employs depthwise separable convolutions to efficiently extract spatio–temporal features while providing preliminary interpretability via feature‐map visualization. Subsequently, temporal convolutional networks (TCNs)~\cite{bai2018empirical,lin2019medical} leveraging dilated causal convolutions achieved parallelizable computation and extended receptive fields, surpassing LSTM‐based approaches~\cite{zhou2016attention,shen2016neural,hochreiter1997long} on multiple medical signal classification benchmarks.  Hybrid architectures such as EEG‐Conformer~\cite{song2022eeg} combined convolutional front‐ends with Transformer self‐attention to capture both local and global dependencies and enabled attention‐based interpretability.  More recently, fine-grained Transformer models such as Medformer~\cite{wang2024medformer} introduced cross-channel tokenization and dual-stage self-attention, setting new SOTA accuracy on several public datasets.  The latest MedGNN ~\cite{fan2025medgnn} further augments attention mechanisms with multi‐resolution graph learning to jointly model spatial multi‐scale channel dependencies and temporal dynamics.


\textbf{Model-centric Transformer-based time series methods.}
In time series analysis, Transformer-based models learn complex dependencies through diversified architectural designs: the vanilla Transformer~\cite{vaswani2017attention} first introduced multi-head self-attention and sinusoidal positional encoding to model temporal correlations globally; Informer~\cite{zhou2021informer} employs ProbSparse attention to select key time steps and compress sequence length, thereby reducing the computational cost of long-range dependencies; Reformer~\cite{kitaev2019reformer} incorporates Locality-Sensitive Hashing (LSH) to reduce attention complexity to \(\mathcal{O}(L\log L)\), making it suitable for ultra-long sequences; Autoformer~\cite{wu2021autoformer} proposes an Auto-Correlation mechanism that aggregates periodic subsequences to enhance the implicit capture of cyclic patterns; FEDformer~\cite{zhou2022fedformer} performs seasonal–trend decomposition in the frequency domain and uses compressed Fourier coefficients to enable cross-frequency attention interactions; Crossformer~\cite{zhang2022crossformer} designs a two-stage attention mechanism across time and feature dimensions to implicitly fuse multivariate spatiotemporal couplings; iTransformer~\cite{liu2023itransformer} innovatively treats time steps as channel dimensions and applies standard attention to implicitly learn nonlinear inter-variable relationships; PatchTST~\cite{nie2022time} segments continuous time steps into patch-based tokens and uses a combination of local and global attention to capture multi-scale temporal patterns; Medformer~\cite{wang2024medformer} introduces multi-granularity patch embeddings and cross-channel attention for medical signals, implicitly modeling the heterogeneous couplings of physiological metrics; and MedGNN~\cite{fan2025medgnn} combines graph attention with frequency-differential networks to incorporate medical topological priors into implicit spatiotemporal dependency learning.

\section{Method}

\subsection{Channel-Imposed Fusion}
\begin{figure}[htbp]
    \centering
 \includegraphics[
        width=\textwidth,
        height=0.7\textheight,  
        keepaspectratio,        
        ]{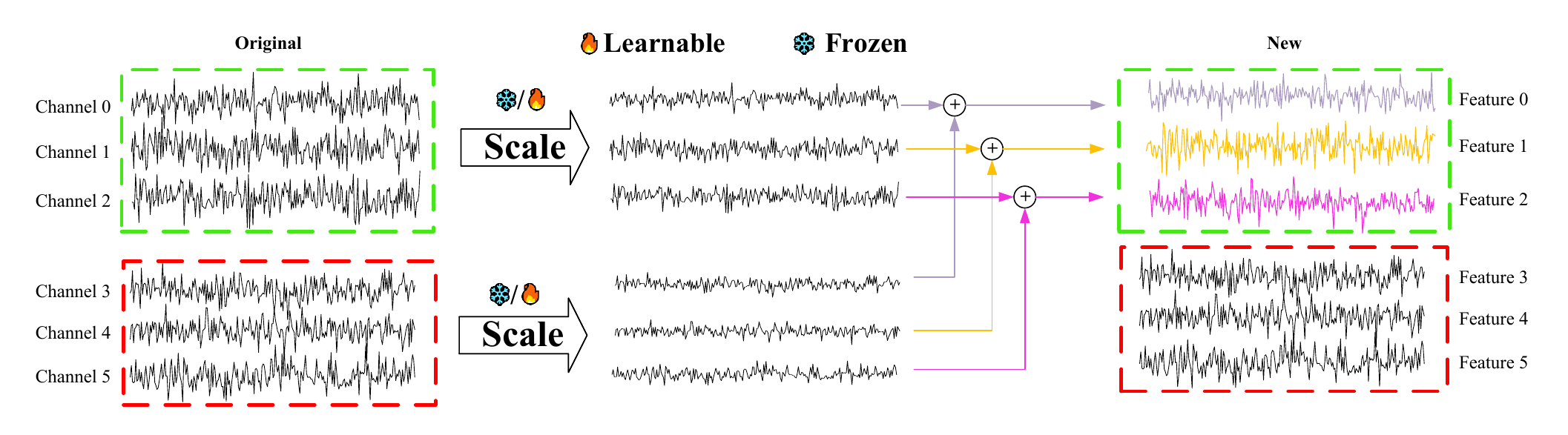}
    \caption{The implementation process of the Channel-Imposed Fusion method.}
    \label{fig:arc_cif}
\end{figure}

In this section, we analyze the data fusion process using Singular Value Decomposition (SVD)~\cite{golub1971singular,klema1980singular, harner1990singular,1997Spatial}. Suppose the original time-series data matrix \( X \) has dimensions \( \mathbb{R}^{T \times C} \), where \( T \) represents the temporal length and \( C \) denotes the number of channels. As shown in Fig.\ref{fig:arc_cif}, we partition \( X \) along the channel dimension into two submatrices: the first \( n \) channels form \( X_{i} = X[:, :n] \), while the last \( n \) channels form \( X_{j} = X[:, -n:] \). The goal of fusion is to combine these two submatrices into a new representation:  
\begin{equation}
X_{fused} = a X_{i} + b X_{j},
\end{equation}
where \( a \) and \( b \) are learnable parameters that control the contribution of the front and back segments in the fused matrix. To understand this fusion process, we first apply SVD to both submatrices:
\begin{equation}
X_{i} = U_1 \Sigma_1 V_1^T, \quad X_{j} = U_2 \Sigma_2 V_2^T,
\end{equation}
where \( U_1, U_2 \in \mathbb{R}^{T \times T} \) are the left singular vectors representing temporal patterns, \( \Sigma_1, \Sigma_2 \in \mathbb{R}^{T \times n} \) are diagonal matrices containing the singular values that indicate the importance of each channel, and \( V_1, V_2 \in \mathbb{R}^{n \times n} \) are the right singular vectors capturing channel relationships. The fused matrix \( X_{\text{fused}} \) is then constructed by linearly combining these two matrices. Depending on the degree of similarity between \( X_{\text{i}} \) and \( X_{\text{j}} \), the fusion process either reduces redundancy or increases data diversity.  
When the correlation between \( X_{i} \) and \( X_{j} \) is high, their singular values and left singular vectors \( U_1 \) and \( U_2 \) are similar. This implies that both matrices capture similar temporal patterns, and the fused matrix \( X_{fused} \) primarily retains these common structures. Mathematically, if \( U_1 \approx U_2 \), the fused matrix can be approximated as:
\begin{equation}
X_{\text{fused}} \approx U_1 \big(a \Sigma_1 + b \Sigma_2 \big) V_1^T.
\end{equation}

In this case, \( X_{fused} \) is largely determined by the shared temporal patterns \( U_1 \), while the weighted sum of \( \Sigma_1 \) and \( \Sigma_2 \) reflects the contribution of each channel. Since the temporal patterns are similar, the fusion process does not introduce significant new information, effectively reducing redundancy.  
On the other hand, when \( X_{i} \) and \( X_{j} \) have low correlation, their singular values and left singular vectors \( U_1 \) and \( U_2 \) differ significantly. This means that the two matrices capture distinct temporal patterns, and the fused matrix \( X_{fused} \) will combine these different components, enhancing data diversity. Mathematically, if \( U_1 \) and \( U_2 \) are dissimilar, the fused matrix can be expressed as:
\begin{equation}
X_{fused} = a U_1 \Sigma_1 V_1^T + b U_2 \Sigma_2 V_2^T.
\end{equation}

In this scenario, the fused matrix incorporates both the temporal patterns from \( U_1 \) and \( U_2 \), effectively combining the complementary information from both matrices. This leads to an increase in data diversity, as each matrix contributes distinct temporal patterns and channel relationships to the final fused representation. Thus, the fusion process allows for either redundancy reduction by preserving shared patterns or diversity enhancement by integrating complementary information, depending on the correlation between the front and back segments.

It is worth noting that no matter how the $n$ channels are selected for partitioning and fusion, this operation is meaningful. Each group of channels contains a portion of the time-series information, and by performing SVD decomposition and linear combination on different channel subsets, one can explore their potential similarities and complementarities, thereby achieving either redundancy reduction or diversity enhancement. Therefore, this partitioning strategy is generally applicable and is not limited to the first $n$ and the last $n$ channels.

\textbf{Optimizing SNR via CIF:} Let two observed signals be \( x_1 = s_1 + \epsilon_1 \) and \( x_2 = s_2 + \epsilon_2 \), where \( s_1, s_2 \) are zero-mean useful signals with variance \( \sigma_s^2 \) and correlation coefficient \( \rho \), and \( \epsilon_1, \epsilon_2 \) are zero-mean noise components with variance \( \sigma_\epsilon^2 \) and correlation coefficient \( \gamma \). Assume that the signal and noise components are mutually uncorrelated. When applying the CIF operation via a linear combination \( y = a x_1 + b x_2 \), the output signal-to-noise ratio (SNR) becomes (details in Appendix~\ref{prove_CIF}):
\begin{equation}
\mathrm{SNR}_{\text{out}} = \mathrm{SNR}_{\text{in}} \cdot \frac{a^2 + b^2 + 2ab\,\rho}{a^2 + b^2 + 2ab\,\gamma}.
\label{eq:snr_gain}
\end{equation}

where \(\mathrm{SNR}_{\mathrm{in}} = \sigma_s^2 / \sigma_\epsilon^2\). When the numerator exceeds the denominator (i.e., 
\(\frac{a^2 + b^2 + 2ab\rho}{a^2 + b^2 + 2ab\gamma} > 1\)),
the SNR improves. This phenomenon occurs in two distinct modes: the \emph{Difference Mode} (\(ab < 0\) and \(\rho < \gamma\)) and the \emph{Cooperative Mode} (\(ab > 0\) and \(\rho > \gamma\)). In the Difference Mode, the correlation of noise sources (e.g., ocular artifacts) is higher than that of the signal, enhancing the SNR by suppressing correlated noise. In the Cooperative Mode, task-related brain regions exhibit synchronized activity while noise remains uncorrelated, leading to an increase in SNR by constructively accumulating and amplifying task-relevant signals. 
Both modes optimize the SNR by adjusting parameters $a$ and $b$, showcasing CIF's ability to suppress noise and enhance the signal under varying correlations.

\subsection{HM-BiTCN Structure Design and Theoretical Analysis}
\begin{figure}[htbp]
    \centering
    \includegraphics[width=1\textwidth]{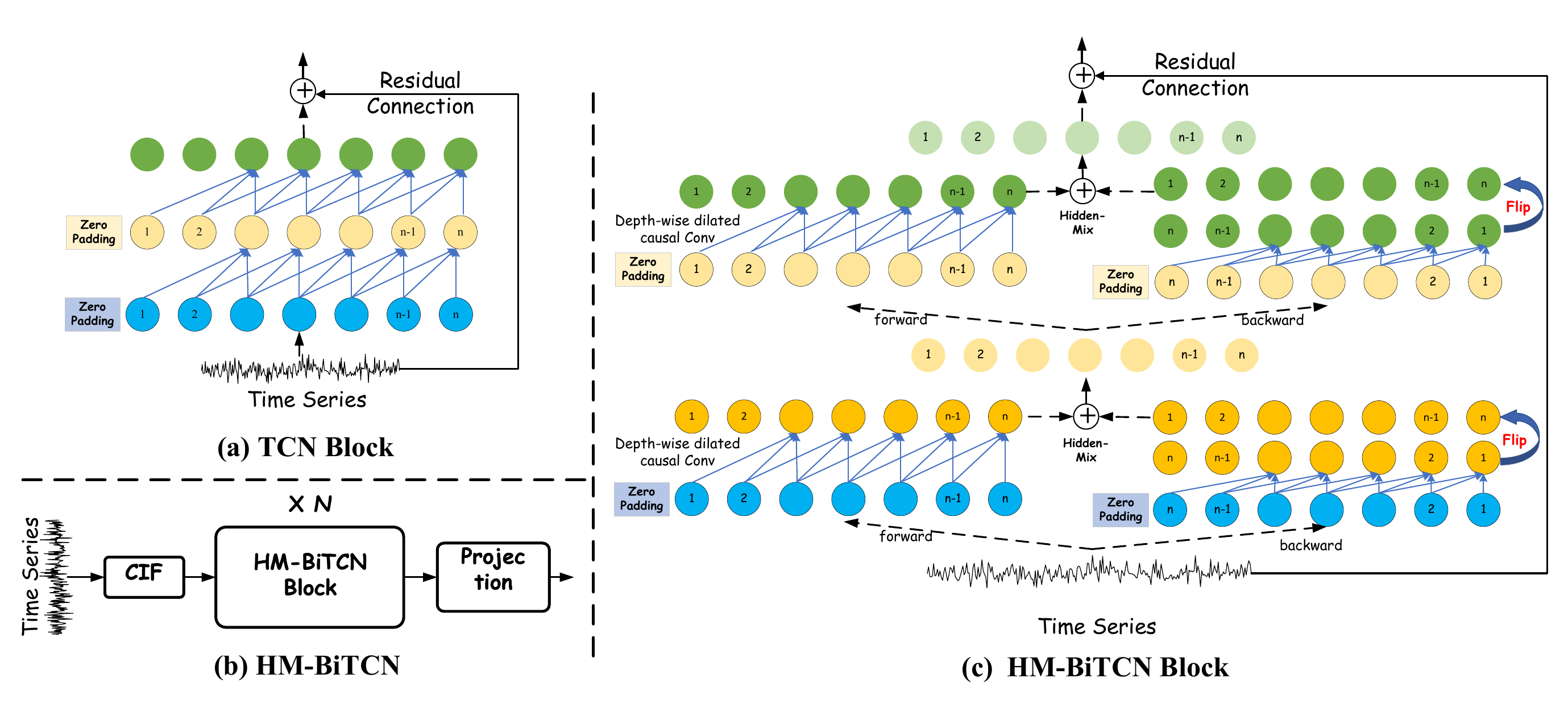}

    \caption{HM-BiTCN Architecture Diagram.}
    \label{fig:HM-BiTCN}
\end{figure}

To demonstrate the value of data-centric approaches and to show that simple models can also achieve strong performance, as shown in Figure\ref{fig:HM-BiTCN}~(a), the conventional TCN only considers unidirectional causal relationships in time series. In contrast, as shown in Figure\ref{fig:HM-BiTCN}~(c), our proposed HM-BiTCN simply introduces bidirectional relationships and performs feature mixing at each layer. This design not only further enhances the signals with improved SNR through CIF but also captures both forward and backward causal structures, thereby making fuller use of the information contained in the data. See Appendix~\ref{app:hmbitcn} and Appendix~\ref{app:Pseudocode} for details.
\textbf{We must emphasize that such a simple modification does not constitute an innovation in itself. Our motivation is to demonstrate that simple models, when combined with a data-centric approach, can also surpass existing SOTA models.}

\section{Experiments}

\textbf{Medical Time Series Datasets.}
(1) \textbf{\textsc{APAVA}}~\cite{escudero2006analysis} is an EEG dataset where each sample is assigned a binary label indicating whether the subject has Alzheimer’s disease. (2) \textbf{\textsc{TDBrain}}~\cite{van2022two} is an EEG dataset with a binary label assigned to each sample, indicating whether the subject has Parkinson’s disease. (3) \textbf{\textsc{ADFTD}}~\cite{miltiadous2023dataset,miltiadous2023dice} is an EEG dataset with a three-class label for each sample, categorizing the subject as Healthy, having Frontotemporal Dementia, or Alzheimer’s disease. (4) \textbf{\textsc{PTB}}~\cite{physiobank2000physionet} is an ECG dataset where each sample is labeled with a binary indicator of Myocardial Infarction. (5) \textbf{\textsc{PTB-XL}}~\cite{wagner2020ptb} is an ECG dataset with a five-class label for each sample, representing various heart conditions.

\textbf{Baselines.}
We compare with 12 state-of-the-art time series transformer methods: Autoformer~\cite{wu2021autoformer}, Crossformer~\cite{zhang2022crossformer}, FEDformer~\cite{zhou2022fedformer}, Informer~\cite{zhou2021informer}, iTransformer~\cite{liu2023itransformer}, MTST~\cite{zhang2024multi}, Nonformer~\cite{liu2022non}, PatchTST~\cite{nie2022time}, Reformer~\cite{kitaev2019reformer}, vanilla Transformer~\cite{vaswani2017attention}, Medformer~\cite{wang2024medformer}, MedGNN~\cite{fan2025medgnn}.

\textbf{Implementation.}
We employ six evaluation metrics: accuracy, precision (macro-averaged), recall (macro-averaged), F1 score (macro-averaged), AUROC (macro-averaged), and AUPRC (macro-averaged). The training process is conducted with five random seeds (41-45) on fixed training, validation, and test sets to compute the mean and standard deviation of the models. All experiments were conducted using an NVIDIA RTX 3090 GPU and implemented with PyTorch version 1.11.0~\cite{paszke2017automatic}. 
We consider two dataset partitioning strategies: (i) \textbf{Subject-Dependent Split}, where the dataset is split at the sample level such that samples from the same subject may appear in both training and test sets, which can cause information leakage and yield overly optimistic performance estimates; and (ii) \textbf{Subject-Independent Split}, where the dataset is split at the subject level, ensuring that each subject appears only in one of the train, validation, or test sets, simulating real-world diagnostic scenarios but introducing challenges due to inter-subject variability.


\subsection{Results of Subject-Dependent}
\label{sub:subject_dependent_result}

\begin{table}[h]
\centering
\def\arraystretch{1.0}
\caption{\textbf{Results of Subject-Dependent Setup.} 
Results of the ADFTD dataset under this setup are presented here.The best result is highlighted in \textbf{bold}, and the second-best is \underline{underlined}.
}
\label{tab:subject-dependent-report}
\resizebox{\textwidth}{!}{%
\begin{tabular}{clcccccc}
    \toprule
      \textbf{Datasets} & \textbf{Models} & \multicolumn{1}{c}{\textbf{Accuracy $\uparrow$}} & \multicolumn{1}{c}{\textbf{Precision $\uparrow$}} & \multicolumn{1}{c}{\textbf{Recall $\uparrow$}} & \multicolumn{1}{c}{\textbf{F1 score $\uparrow$}} & \multicolumn{1}{c}{\textbf{AUROC $\uparrow$}} & \multicolumn{1}{c}{\textbf{AUPRC $\uparrow$}} \\
    \midrule
    \multirow{11}{*}{\begin{tabular}[c]{@{}l@{}}\;\makecell{\textbf{ADFTD} \\ (3-Classes) \\ Reported} \end{tabular}} 
    & \textbf{Autoformer}~\cite{wu2021autoformer} & 87.83\std{1.62} & 87.63\std{1.66} & 87.22\std{1.97} & 87.38\std{1.79} & 96.59\std{0.88} & 93.82\std{1.64} \\
    & \textbf{Crossformer}~\cite{zhang2022crossformer} & 89.35\std{1.32} & 89.00\std{1.44} & 88.79\std{1.37} & 88.88\std{1.40} & 97.52\std{0.58} & 95.45\std{1.03} \\
    & \textbf{FEDformer}~\cite{zhou2022fedformer} & 77.63\std{2.37} & 76.76\std{2.17} & 76.68\std{2.48} & 76.60\std{2.46} & 91.67\std{1.34} & 84.94\std{2.11} \\
    & \textbf{Informer}~\cite{zhou2021informer} & 90.93\std{0.90} & 90.74\std{0.71} & 90.50\std{1.14} & 90.60\std{0.94} & 98.19\std{0.27} & 96.51\std{0.49} \\
    & \textbf{iTransformer}~\cite{liu2023itransformer} & 64.90\std{0.25} & 62.53\std{0.27} & 62.21\std{0.26} & 62.25\std{0.33} & 81.52\std{0.29} & 68.87\std{0.49} \\
    & \textbf{MTST}~\cite{zhang2024multi}  & 65.08\std{0.69} & 63.85\std{0.80} & 62.71\std{0.64} & 63.03\std{0.58} & 81.36\std{0.56} & 69.34\std{0.89} \\
    & \textbf{Nonformer}~\cite{liu2022non}  & 96.12\std{0.47} & 95.94\std{0.56} & 95.99\std{0.38} & 95.96\std{0.47} & 99.59\std{0.09} & 99.08\std{0.16} \\
    & \textbf{PatchTST}~\cite{nie2022time}  & 66.26\std{0.40} & 65.08\std{0.41} & 64.97\std{0.51} & 64.95\std{0.42} & 83.07\std{0.45} & 71.70\std{0.61} \\
    & \textbf{Reformer}~\cite{kitaev2019reformer}  & 91.51\std{1.75} & 91.15\std{1.79} & 91.65\std{1.56} & 91.14\std{1.83} & 98.85\std{0.35} & 97.88\std{0.60} \\
    & \textbf{Transformer}~\cite{vaswani2017attention}  & 97.00\std{0.43} & 96.87\std{0.53} & 96.86\std{0.36} & 96.86\std{0.44} & 99.75\std{0.04} & 99.42\std{0.07} \\
    & \textbf{Medformer}~\cite{wang2024medformer}  & 97.62\std{0.34} & 97.53\std{0.33} & 97.48\std{0.40} & 97.50\std{0.36} & 99.83\std{0.05} & 99.62\std{0.12} \\

    & \textbf{MedGNN}~\cite{fan2025medgnn}  & 98.42\std{0.04} & 98.31\std{0.02} & 98.29\std{0.05} & 98.30\std{0.12} & 99.93\std{0.11} & - \\

    & \textbf{Medformer + CIF}  & \underline{98.87\std{0.26}}& \underline{98.77\std{0.27}}& \underline{98.86\std{0.27}}& \underline{98.81\std{0.27}}& \underline{99.96\std{0.01}}& \underline{99.92\std{0.03}} \\
& \textbf{MedGNN + CIF} & \textbf{99.60\std{0.09}}& \textbf{99.60\std{0.11}}& \textbf{99.58\std{0.09}}& \textbf{99.59\std{0.10}}& \textbf{99.99\std{0.01}}& \textbf{99.97\std{0.01}} \\
\bottomrule
\end{tabular}
}

\end{table}

We reproduced 12 baselines. Table~\ref{tab:subject-dependent-report} lists their reported results.
Experimental results show that integrating the CIF method into MedGNN and Medformer outperforms existing approaches, fully demonstrating its effectiveness and superiority.

\subsection{Results of Subject-Independent}
\label{sub:subject_independent_result}
 \begin{table}[h]
\centering
\def\arraystretch{0.95}
\caption{\textbf{Results of Subject-Independent Setup.} 
The results we compare include those reported by Medforme~\cite{wang2024medformer} and MedGNN~\cite{fan2025medgnn}. The best result is highlighted in \textbf{bold}, and the second-best is \underline{underlined}.
}
\label{tab:subject-independent-report}
\resizebox{\textwidth}{!}{%
\begin{tabular}{clcccccc}
    \toprule
      \textbf{Datasets} & \textbf{Models} & \multicolumn{1}{c}{\textbf{Accuracy $\uparrow$}} & \multicolumn{1}{c}{\textbf{Precision $\uparrow$}} & \multicolumn{1}{c}{\textbf{Recall $\uparrow$}} & \multicolumn{1}{c}{\textbf{F1 score $\uparrow$}} & \multicolumn{1}{c}{\textbf{AUROC $\uparrow$}} & \multicolumn{1}{c}{\textbf{AUPRC $\uparrow$}} \\
    \midrule
    \multirow{11}{*}{\begin{tabular}[c]{@{}l@{}}\;\makecell{\textbf{APAVA} \\ (2-Classes) \\Reported} \end{tabular}} 
    & \textbf{Autoformer}~\cite{wu2021autoformer} &    68.64\std{1.82} & 68.48\std{2.10} & 68.77\std{2.27} & 68.06\std{1.94} & 75.94\std{3.61} & 74.38\std{4.05} \\
    & \textbf{Crossformer}~\cite{zhang2022crossformer} & 73.77\std{1.95} & 79.29\std{4.36} & 68.86\std{1.70} & 68.93\std{1.85} & 72.39\std{3.33} & 72.05\std{3.65} \\
    & \textbf{FEDformer}~\cite{zhou2022fedformer} & 74.94\std{2.15} & 74.59\std{1.50} & 73.56\std{3.55} & 73.51\std{3.39} & 83.72\std{1.97} & 82.94\std{2.37}  \\
    & \textbf{Informer}~\cite{zhou2021informer} & 73.11\std{4.40} & 75.17\std{6.06} & 69.17\std{4.56} & 69.47\std{5.06} & 70.46\std{4.91} & 70.75\std{5.27}  \\
    & \textbf{iTransformer}~\cite{liu2023itransformer} & 74.55\std{1.66} & 74.77\std{2.10} & 71.76\std{1.72} & 72.30\std{1.79} & 85.59\std{1.55} & \underline{84.39\std{1.57}} \\
    & \textbf{MTST}~\cite{zhang2024multi}  & 71.14\std{1.59} & 79.30\std{0.97} & 65.27\std{2.28} & 64.01\std{3.16} & 68.87\std{2.34} & 71.06\std{1.60} \\
    & \textbf{Nonformer}~\cite{liu2022non}  & 71.89\std{3.81} & 71.80\std{4.58} & 69.44\std{3.56} & 69.74\std{3.84} & 70.55\std{2.96} & 70.78\std{4.08}  \\
    & \textbf{PatchTST}~\cite{nie2022time}   & 67.03\std{1.65} & 78.76\std{1.28} & 59.91\std{2.02} & 55.97\std{3.10} & 65.65\std{0.28} & 67.99\std{0.76}  \\
    & \textbf{Reformer}~\cite{kitaev2019reformer}  & 78.70\std{2.00} & 82.50\std{3.95} & 75.00\std{1.61} & 75.93\std{1.82} & 73.94\std{1.40} & 76.04\std{1.14}  \\
    & \textbf{Transformer}~\cite{vaswani2017attention}  & 76.30\std{4.72} & 77.64\std{5.95} & 73.09\std{5.01} & 73.75\std{5.38} & 72.50\std{6.60} & 73.23\std{7.60} \\
    & \textbf{Medformer}~\cite{wang2024medformer} & 78.74\std{0.64} & 81.11\std{0.84} & 75.40\std{0.66} & 76.31\std{0.71} & 83.20\std{0.91} & 83.66\std{0.92} \\
  
     & \textbf{MedGNN}~\cite{fan2025medgnn} & \underline{82.60\std{0.35}} & \textbf{87.70\std{0.22}} & \underline{78.93\std{0.09}} & \underline{80.25\std{0.16}} & \underline{85.93\std{0.26}} & - \\
     & \textbf{HM-BiTCN + CIF} & \textbf{85.16\std{1.55}} & \underline{84.76\std{1.62}} & \textbf{85.33\std{1.27}} & \textbf{84.82\std{1.49}} & \textbf{94.06\std{1.07}} & \textbf{94.21\std{0.99}}  \\

    \midrule
    \multirow{11}{*}{\begin{tabular}[c]{@{}l@{}}\;\makecell{\textbf{TDBrain} \\ (2-Classes) \\Reported} \end{tabular}} 
    & \textbf{Autoformer}~\cite{wu2021autoformer} & 87.33\std{3.79} & 88.06\std{3.56} & 87.33\std{3.79} & 87.26\std{3.84} & 93.81\std{2.26} & 93.32\std{2.42}  \\
    & \textbf{Crossformer}~\cite{zhang2022crossformer} & 81.56\std{2.19} & 81.97\std{2.25} & 81.56\std{2.19} & 81.50\std{2.20} & 91.20\std{1.78} & 91.51\std{1.71}  \\
    & \textbf{FEDformer}~\cite{zhou2022fedformer} & 78.13\std{1.98} & 78.52\std{1.91} & 78.13\std{1.98} & 78.04\std{2.01} & 86.56\std{1.86} & 86.48\std{1.99}  \\
    & \textbf{Informer}~\cite{zhou2021informer} & 89.02\std{2.50} & 89.43\std{2.14} & 89.02\std{2.50} & 88.98\std{2.54} & 96.64\std{0.68} & \underline{96.75\std{0.63}}  \\
    & \textbf{iTransformer}~\cite{liu2023itransformer} & 74.67\std{1.06} & 74.71\std{1.06} & 74.67\std{1.06} & 74.65\std{1.06} & 83.37\std{1.14} & 83.73\std{1.27}  \\
    & \textbf{MTST}~\cite{zhang2024multi}  & 76.96\std{3.76} & 77.24\std{3.59} & 76.96\std{3.76} & 76.88\std{3.83} & 85.27\std{4.46} & 82.81\std{5.64}  \\
    & \textbf{Nonformer}~\cite{liu2022non}  & 87.88\std{2.48} & 88.86\std{1.84} & 87.88\std{2.48} & 87.78\std{2.56} & 97.05\std{0.68} & 96.99\std{0.68} \\
    & \textbf{PatchTST}~\cite{nie2022time}   & 79.25\std{3.79} & 79.60\std{4.09} & 79.25\std{3.79} & 79.20\std{3.77} & 87.95\std{4.96} & 86.36\std{6.67} \\
    & \textbf{Reformer}~\cite{kitaev2019reformer}  & 87.92\std{2.01} & 88.64\std{1.40} & 87.92\std{2.01} & 87.85\std{2.08} & 96.30\std{0.54} & 96.40\std{0.45}  \\
    & \textbf{Transformer}~\cite{vaswani2017attention}  & 87.17\std{1.67} & 87.99\std{1.68} & 87.17\std{1.67} & 87.10\std{1.68} & 96.28\std{0.92} & 96.34\std{0.81}  \\
    & \textbf{Medformer }~\cite{wang2024medformer}  & 89.62\std{0.81} & 89.68\std{0.78} & 89.62\std{0.81} & 89.62\std{0.81} & 96.41\std{0.35} & 96.51\std{0.33}  \\
     & \textbf{MedGNN }~\cite{fan2025medgnn}  & \underline{91.04\std{0.09}} & \underline{91.15\std{0.12}} & \underline{91.04\std{0.20}} & \underline{91.04\std{0.08}} & \underline{96.74\std{0.04}} & - \\

     & \textbf{HM-BiTCN + CIF}  & \textbf{93.13\std{1.41}} & \textbf{93.33\std{1.37}} & \textbf{93.13\std{1.41}} & \textbf{93.12\std{1.42}} & \textbf{98.62\std{0.66}} & \textbf{98.68\std{0.63}}\\
    \midrule
    \multirow{11}{*}{\begin{tabular}[c]{@{}l@{}}\;\makecell{\textbf{ADFTD} \\ (3-Classes) \\Reported} \end{tabular}} 
    & \textbf{Autoformer}~\cite{wu2021autoformer} & 45.25\std{1.48} & 43.67\std{1.94} & 42.96\std{2.03} & 42.59\std{1.85} & 61.02\std{1.82} & 43.10\std{2.30}  \\
    & \textbf{Crossformer}~\cite{zhang2022crossformer} & 50.45\std{2.31} & 45.57\std{1.63} & 45.88\std{1.82} & 45.50\std{1.70} & 66.45\std{2.03} & 48.33\std{2.05}  \\
    & \textbf{FEDformer}~\cite{zhou2022fedformer} & 46.30\std{0.59} & 46.05\std{0.76} & 44.22\std{1.38} & 43.91\std{1.37} & 62.62\std{1.75} & 46.11\std{1.44}  \\
    & \textbf{Informer}~\cite{zhou2021informer} & 48.45\std{1.96} & 46.54\std{1.68} & 46.06\std{1.84} & 45.74\std{1.38} & 65.87\std{1.27} & 47.60\std{1.30}  \\
    & \textbf{iTransformer}~\cite{liu2023itransformer} & 52.60\std{1.59} & 46.79\std{1.27} & 47.28\std{1.29} & 46.79\std{1.13} & 67.26\std{1.16} & 49.53\std{1.21} \\
    & \textbf{MTST}~\cite{zhang2024multi}  & 45.60\std{2.03} & 44.70\std{1.33} & 45.05\std{1.30} & 44.31\std{1.74} & 62.50\std{0.81} & 45.16\std{0.85}  \\
    & \textbf{Nonformer}~\cite{liu2022non}  & 49.95\std{1.05} & 47.71\std{0.97} & 47.46\std{1.50} & 46.96\std{1.35} & 66.23\std{1.37} & 47.33\std{1.78}  \\
    & \textbf{PatchTST}~\cite{nie2022time}   & 44.37\std{0.95} & 42.40\std{1.13} & 42.06\std{1.48} & 41.97\std{1.37} & 60.08\std{1.50} & 42.49\std{1.79}  \\
    & \textbf{Reformer}~\cite{kitaev2019reformer}  & 50.78\std{1.17} & 49.64\std{1.49} & 49.89\std{1.67} & 47.94\std{0.69} & 69.17\std{1.58} & \underline{51.73\std{1.94}} \\
    & \textbf{Transformer}~\cite{vaswani2017attention}  & 50.47\std{2.14} & 49.13\std{1.83} & 48.01\std{1.53} & 48.09\std{1.59} & 67.93\std{1.59} & 48.93\std{2.02} \\
    & \textbf{Medformer }~\cite{wang2024medformer}  & 53.27\std{1.54} & 51.02\std{1.57} & 50.71\std{1.55} & 50.65\std{1.51} & 70.93\std{1.19} & 51.21\std{1.32} \\
    & \textbf{MedGNN }~\cite{fan2025medgnn}  & \underline{56.12\std{0.11}} & \underline{55.07\std{0.09}} & \underline{55.47\std{0.34}} & \underline{55.00\std{0.24}} & \underline{74.68\std{0.33}} & -   \\
      & \textbf{HM-BiTCN + CIF}  & \textbf{58.56\std{0.93}} & \textbf{55.65\std{0.81}} & \textbf{55.86\std{0.79}} & \textbf{55.42\std{0.82}} & \textbf{76.07\std{0.59}} & \textbf{59.75\std{0.67}} \\
    \midrule
    \multirow{11}{*}{\begin{tabular}[c]{@{}l@{}}\;\makecell{\textbf{PTB} \\ (2-Classes) \\ Reported} \end{tabular}} 
    & \textbf{Autoformer}~\cite{wu2021autoformer} & 73.35\std{2.10} & 72.11\std{2.89} & 63.24\std{3.17} & 63.69\std{3.84} & 78.54\std{3.48} & 74.25\std{3.53}  \\
    & \textbf{Crossformer}~\cite{zhang2022crossformer} & 80.17\std{3.79} & 85.04\std{1.83} & 71.25\std{6.29} & 72.75\std{7.19} & 88.55\std{3.45} & 87.31\std{3.25}  \\
    & \textbf{FEDformer}~\cite{zhou2022fedformer} & 76.05\std{2.54} & 77.58\std{3.61} & 66.10\std{3.55} & 67.14\std{4.37} & 85.93\std{4.31} & 82.59\std{5.42}  \\
    & \textbf{Informer}~\cite{zhou2021informer} & 78.69\std{1.68} & 82.87\std{1.02} & 69.19\std{2.90} & 70.84\std{3.47} & 92.09\std{0.53} & 90.02\std{0.60}  \\
    & \textbf{iTransformer}~\cite{liu2023itransformer} & 83.89\std{0.71} & \underline{88.25\std{1.18}} & 76.39\std{1.01} & 79.06\std{1.06} & 91.18\std{1.16} & \underline{90.93\std{0.98}}  \\
    & \textbf{MTST}~\cite{zhang2024multi}  & 76.59\std{1.90} & 79.88\std{1.90} & 66.31\std{2.95} & 67.38\std{3.71} & 86.86\std{2.75} & 83.75\std{2.84}  \\
    & \textbf{Nonformer}~\cite{liu2022non}  & 78.66\std{0.49} & 82.77\std{0.86} & 69.12\std{0.87} & 70.90\std{1.00} & 89.37\std{2.51} & 86.67\std{2.38}  \\
    & \textbf{PatchTST}~\cite{nie2022time}   & 74.74\std{1.62} & 76.94\std{1.51} & 63.89\std{2.71} & 64.36\std{3.38} & 88.79\std{0.91} & 83.39\std{0.96} \\
    & \textbf{Reformer}~\cite{kitaev2019reformer}  & 77.96\std{2.13} & 81.72\std{1.61} & 68.20\std{3.35} & 69.65\std{3.88} & 91.13\std{0.74} & 88.42\std{1.30} \\
    & \textbf{Transformer}~\cite{vaswani2017attention}  & 77.37\std{1.02} & 81.84\std{0.66} & 67.14\std{1.80} & 68.47\std{2.19} & 90.08\std{1.76} & 87.22\std{1.68}  \\
    & \textbf{Medformer }~\cite{wang2024medformer}  & 83.50\std{2.01} & 85.19\std{0.94} & 77.11\std{3.39} & 79.18\std{3.31} & 92.81\std{1.48} & 90.32\std{1.54}  \\
     & \textbf{MedGNN }~\cite{fan2025medgnn}  & \underline{84.53\std{0.28}} & 87.35\std{0.45} & \underline{77.90\std{0.66}} & \underline{80.40\std{0.62}} & \underline{93.31\std{0.46}} & - \\

     & \textbf{HM-BiTCN + CIF}  & \textbf{88.29\std{1.45}} & \textbf{90.66\std{1.48}} & \textbf{83.21\std{2.02}} & \textbf{85.59\std{1.96}} & \textbf{94.28\std{0.93}} & \textbf{93.78\std{1.11}}  \\

    \midrule
    \multirow{11}{*}{\begin{tabular}[c]{@{}l@{}}\;\makecell{\textbf{PTB-XL} \\ (5-Classes) \\ Reported} \end{tabular}} 
    & \textbf{Autoformer}~\cite{wu2021autoformer} & 61.68\std{2.72} & 51.60\std{1.64} & 49.10\std{1.52} & 48.85\std{2.27} & 82.04\std{1.44} & 51.93\std{1.71}  \\
    & \textbf{Crossformer}~\cite{zhang2022crossformer} & 73.30\std{0.14} & 65.06\std{0.35} & \textbf{61.23\std{0.33}} & \underline{62.59\std{0.14}} & 90.02\std{0.06} & \underline{67.43\std{0.22}} \\
    & \textbf{FEDformer}~\cite{zhou2022fedformer} & 57.20\std{9.47} & 52.38\std{6.09} & 49.04\std{7.26} & 47.89\std{8.44} & 82.13\std{4.17} & 52.31\std{7.03}  \\
    & \textbf{Informer}~\cite{zhou2021informer} & 71.43\std{0.32} & 62.64\std{0.60} & 59.12\std{0.47} & 60.44\std{0.43} & 88.65\std{0.09} & 64.76\std{0.17}  \\
    & \textbf{iTransformer}~\cite{liu2023itransformer} & 69.28\std{0.22} & 59.59\std{0.45} & 54.62\std{0.18} & 56.20\std{0.19} & 86.71\std{0.10} & 60.27\std{0.21}  \\
    & \textbf{MTST}~\cite{zhang2024multi}  & 72.14\std{0.27} & 63.84\std{0.72} & 60.01\std{0.81} & 61.43\std{0.38} & 88.97\std{0.33} & 65.83\std{0.51}   \\
    & \textbf{Nonformer}~\cite{liu2022non}  & 70.56\std{0.55} & 61.57\std{0.66} & 57.75\std{0.72} & 59.10\std{0.66} & 88.32\std{0.36} & 63.40\std{0.79} \\
    & \textbf{PatchTST}~\cite{nie2022time}   & 73.23\std{0.25} & \underline{65.70\std{0.64}} & 60.82\std{0.76} & \textbf{62.61\std{0.34}} & 89.74\std{0.19} & 67.32\std{0.22} \\
    & \textbf{Reformer}~\cite{kitaev2019reformer}  & 71.72\std{0.43} & 63.12\std{1.02} & 59.20\std{0.75} & 60.69\std{0.18} & 88.80\std{0.24} & 64.72\std{0.47}  \\
    & \textbf{Transformer}~\cite{vaswani2017attention}  & 70.59\std{0.44} & 61.57\std{0.65} & 57.62\std{0.35} & 59.05\std{0.25} & 88.21\std{0.16} & 63.36\std{0.29}  \\
    & \textbf{Medformer }~\cite{wang2024medformer} & 72.87\std{0.23} & 64.14\std{0.42} & 60.60\std{0.46} & 62.02\std{0.37} & 89.66\std{0.13} & 66.39\std{0.22}  \\
    & \textbf{MedGNN}~\cite{fan2025medgnn} & \textbf{73.87\std{0.18}} & \textbf{66.26\std{0.29}} & \underline{61.13\std{0.29}} & 62.54\std{0.20} & \underline{90.21\std{0.15}} & - \\
      & \textbf{HM-BiTCN + CIF}  & \underline{73.73\std{0.30}}& 65.41\std{0.67}& 60.70\std{1.08}& 61.89\std{0.91}& \textbf{90.53\std{0.22}}& \textbf{67.75\std{0.75}}\\
    \bottomrule
\end{tabular}
}
\end{table}
Table~\ref{tab:subject-independent-report} presents the results \textbf{reported} by various methods in the subject-independent setting. Our method achieves the highest average scores across six metrics on four out of the five datasets. On PTB-XL, our method tops AUROC and AUPRC and ranks second in Accuracy versus reported results, and ranks first in Accuracy, Precision, AUROC, AUPRC, and second in Recall versus our reproduced results.
Additionally, it is worth noting that in the subject-independent setup, the F1 score of ADFTD is 55.42\%, which is significantly lower than the 99.59\% achieved in the subject-dependent setup. This comparison highlights the challenges of the subject-independent setup, \textbf{which better simulates real-world scenarios}.
\subsubsection{Efficiency Analysis}
\begin{figure}[ht]
    \centering
    \begin{subfigure}{0.49\textwidth}
        \centering
       \includegraphics[width=\linewidth]{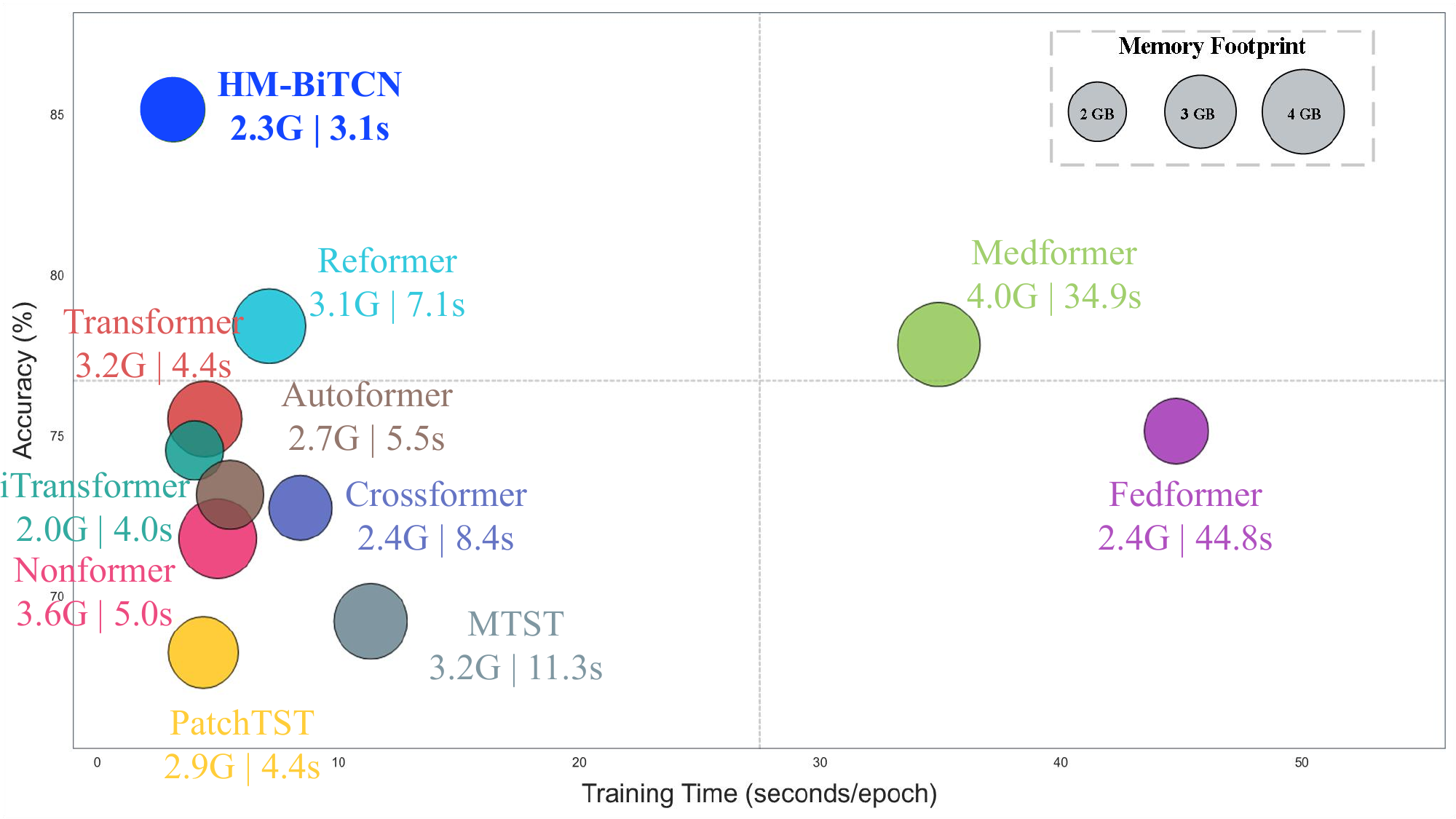}
        \caption{APAVA}  
    \end{subfigure}
    \hfill
    \begin{subfigure}{0.49\textwidth}
        \centering
          \includegraphics[width=\linewidth]{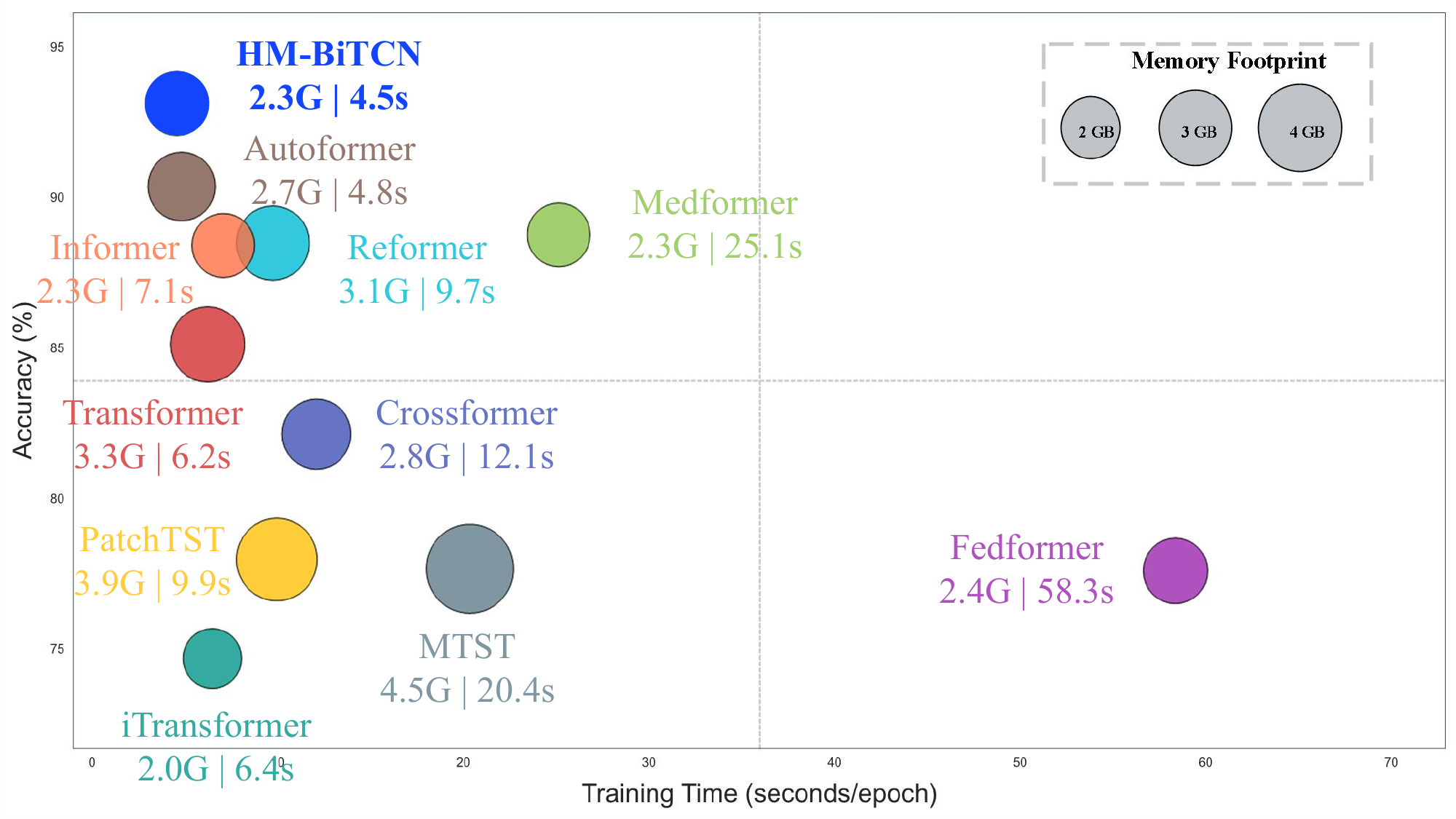}
        \caption{TDBRAIN}  
    \end{subfigure}
    \caption{Effectiveness and efficiency on two datasets (subject-based).} 
        \label{fig_APAVA_TDBrain}
\end{figure}

We evaluate the model efficiency in terms of accuracy, training speed, and memory footprint using two datasets: APAVA and TDBRAIN. In Figure~\ref{fig_APAVA_TDBrain}, a marker closer to the upper-left corner indicates higher accuracy and faster training speed, while a smaller marker area corresponds to lower memory usage. The results show that HM-BiTCN achieves the best overall performance among all baseline methods, demonstrating its high efficiency and reliability across different application scenarios.





\subsection{Ablation Study}
\label{sub:Ablation}

\textbf{(1) Effectiveness of CIF:} Table \ref{tab:abl_differential} demonstrates the excellent performance of combining HM-BiTCN with CIF, confirming the compatibility of the HM-BiTCN with CIF. 
Appendix~\ref{app:ab_hmbiTCN} presents ablation studies on the HM-BiTCN architecture and the performance benefits of integrating CIF into its components.

\begin{table}[!h]
    \centering
    \caption{Exploring the Integration of HM-BiTCN Structure with CIF.}
    \label{tab:abl_differential}
    \scalebox{0.8}{
    \begin{tabular}{l c | c c | c c | c c | c c}
        \toprule[1.5pt]
        \multicolumn{2}{c|}{Datasets} & \multicolumn{2}{c|}{APAVA} & \multicolumn{2}{c|}{ADFTD} & \multicolumn{2}{c|}{PTB} & \multicolumn{2}{c}{TDBRAIN} \\
        \cmidrule(r){1-2} \cmidrule(r){3-4} \cmidrule(r){5-6} \cmidrule(r){7-8} \cmidrule(r){9-10}
        \multicolumn{2}{c|}{Metrics} & Accuracy & F1 Score & Accuracy & F1 Score & Accuracy & F1 Score & Accuracy & F1 Score\\
        \midrule[1pt]
        \multicolumn{2}{c|}{w/ CIF}   & 85.16\std{1.55} & 84.82\std{1.49} & 58.56\std{0.93} & 55.42\std{0.82} & 88.29\std{1.45} & 85.59\std{1.96} & 93.13\std{1.41} & 93.12\std{1.42}  \\
        \multicolumn{2}{c|}{w/o CIF} & 82.49\std{1.40} & 81.60\std{1.39} & 52.05\std{2.22} & 49.48\std{2.70} & 81.87\std{1.87} & 75.84\std{3.20}  & 84.90\std{2.60} & 84.76\std{2.74}\\
        \midrule[1pt]
        \multicolumn{2}{c|}{\textcolor{blue}{\textbf{Improvement}}} 
            & \textcolor{blue}{\textbf{+2.67\%}} & \textcolor{blue}{\textbf{+3.22\%}} 
            & \textcolor{blue}{\textbf{+6.51\%}} & \textcolor{blue}{\textbf{+5.94\%}} 
            & \textcolor{blue}{\textbf{+6.42\%}} & \textcolor{blue}{\textbf{+9.75\%}}
            & \textcolor{blue}{\textbf{+8.23\%}} & \textcolor{blue}{\textbf{+8.36\%}} \\
        \bottomrule[1.5pt]
    \end{tabular}
    }
\end{table}


 
\begin{figure}[ht]
    \centering

    \begin{subfigure}{0.48\textwidth}
        \centering
          \includegraphics[width=\linewidth]{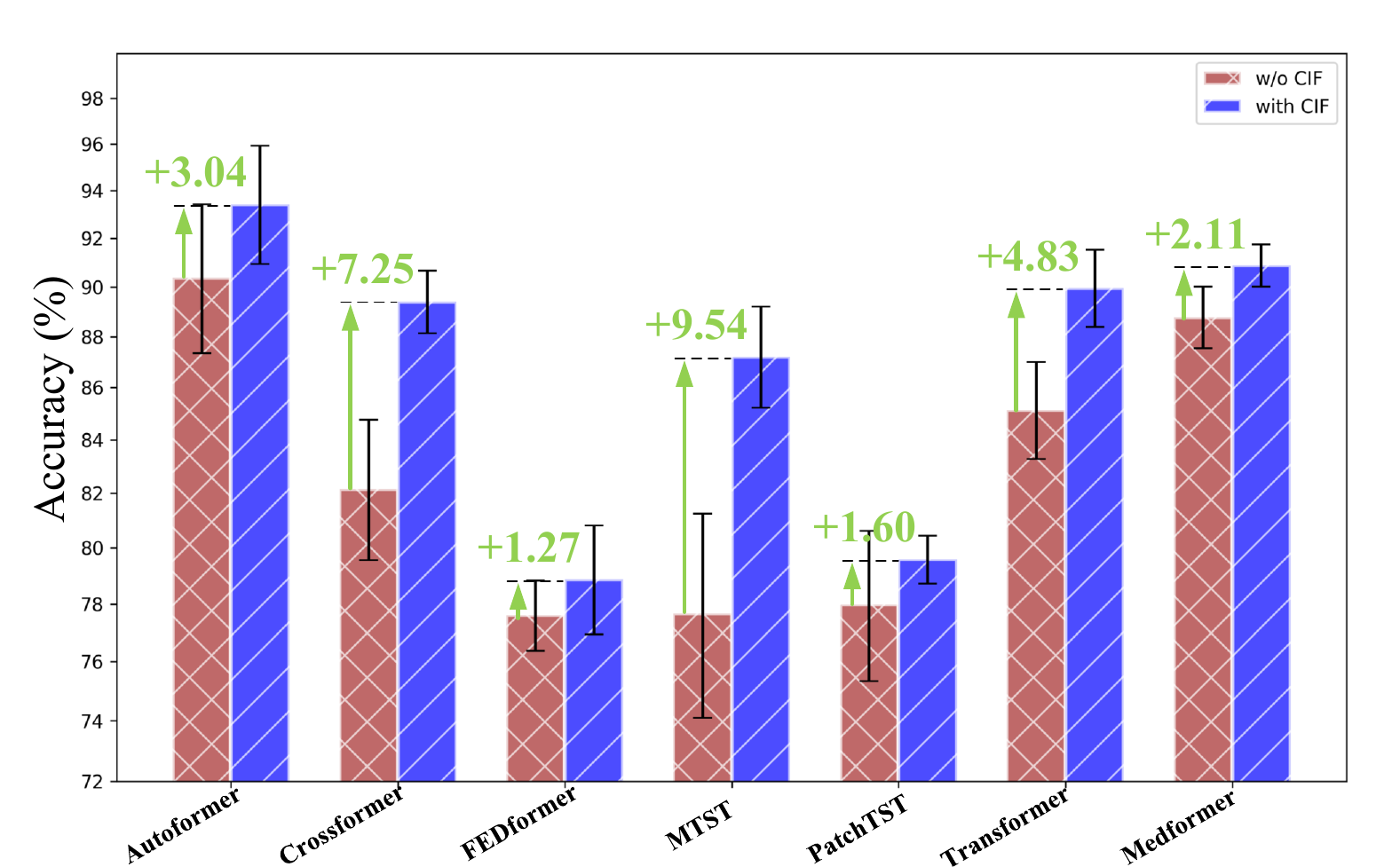}
        \caption{TDBRAIN-Subject}  
    \end{subfigure}
    \hfill
    \begin{subfigure}{0.48\textwidth}
        \centering
          \includegraphics[width=\linewidth]{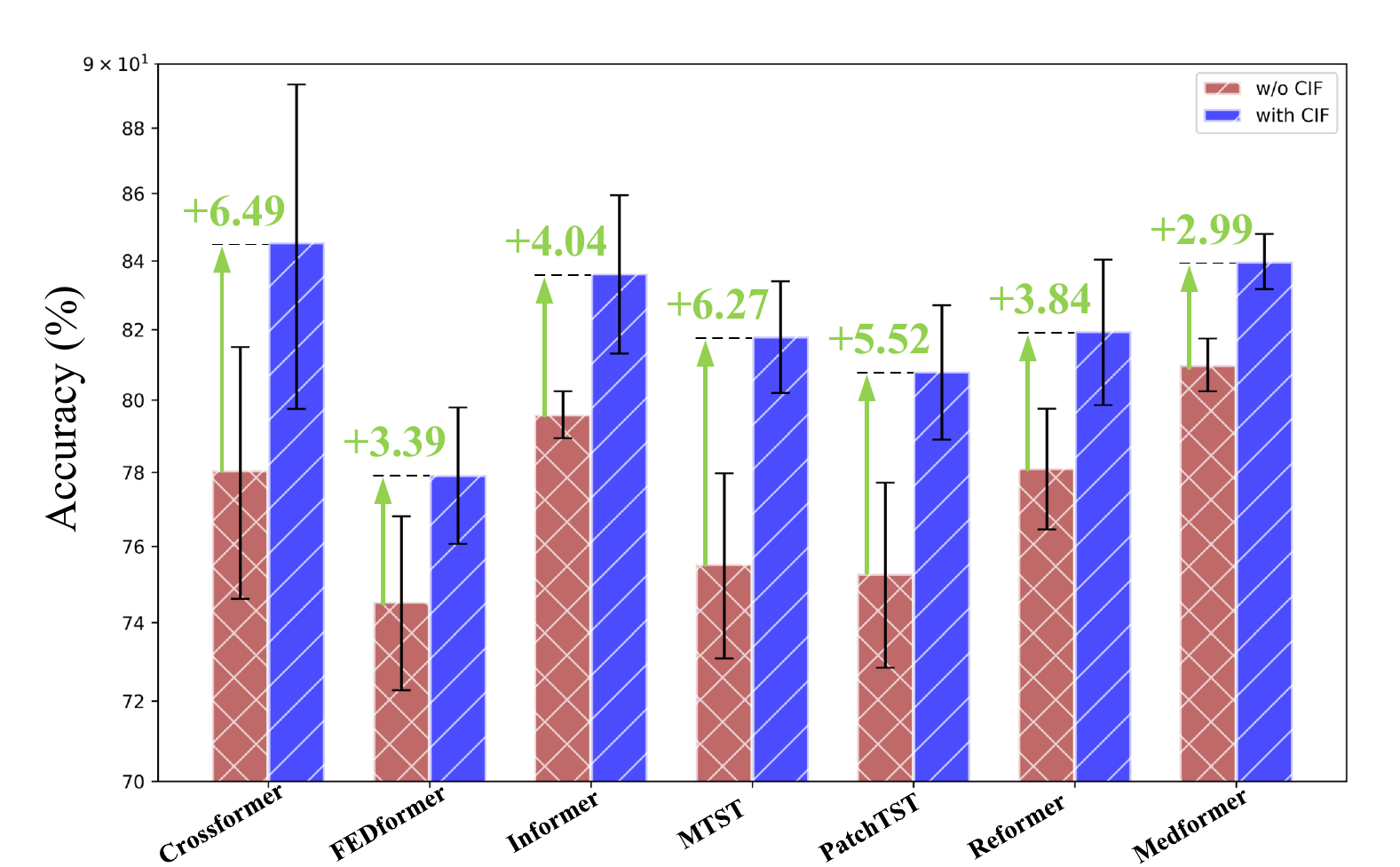}
        \caption{PTB-Subject}  
    \end{subfigure}
    \hfill
 
    \caption{The improvements achieved by various baselines when combined with the CIF method.} 
        \label{fig_APAVA_TDBrain_PTB_APAVA}
\end{figure}

\textbf{(2) Hyperparameter Transfer and Adaptation:} We evaluate the transferability of key hyperparameters (e.g., $a$, $b$, $n$) from HM-BiTCN to other models. If transferred settings underperform, we further fine-tune them for adaptation. 
Figure \ref{fig_APAVA_TDBrain_PTB_APAVA} illustrates the outstanding performance of CIF when combined with other models. 
    
\textbf{(3) Exploring Physiologically-Informed CIF:} In Appendix~\ref{app:PSF}, we further conducted experiments on CIF from the perspective of emphasizing more prominent biological features. The results indicate that incorporating specific biological characteristics can further enhance classification performance.

\textbf{(4)Results on general time series classification tasks}
\begin{table*}[!htbp]
    \centering
     \caption{Performance on the HAR and UCI-HAR non-medical time series datasets. Bold numbers indicate the best results. * denotes the results reported by Medformer.
    }
    \label{tab:human}
    \resizebox{1\textwidth}{!}{
    \begin{tabular}{@{}ll|ccccccccc@{}}
    \toprule
\multicolumn{2}{l|}{\makecell{\textbf{Dataset}  \textbf{/ Metric}}} 
& \makecell{\textbf{Crossformer *} \\ \cite{zhang2022crossformer}} 
& \makecell{\textbf{Reformer *} \\ \cite{kitaev2019reformer}} 
& \makecell{\textbf{Transformer *} \\ \cite{vaswani2017attention}} 
& \makecell{\textbf{TCN *} \\ \cite{bai2018empirical}} 
& \makecell{\textbf{ModernTCN *} \\ \cite{luo2024moderntcn}} 
& \makecell{\textbf{Mamba *} \\ \cite{gu2023mamba}} 
& \makecell{\textbf{Medformer *} \\ \cite{wang2024medformer}} 
& \makecell{\textbf{HM-BiTCN} \\ (This work)} 
& \makecell{\textbf{HM-BiTCN + CIF} \\ (This work)} \\ \midrule

    \multirow{2}{*}{\makecell{\textbf{FLAAP} \\ \textit{(10 Classes)}}}

        & Accuracy & 75.84\std{0.52} & 71.65\std{1.27} & 74.96\std{1.25} & 66.48\std{1.66} 
            & 74.80\std{0.96} & 64.87\std{2.78} & 76.44\std{0.64} & 76.08\std{0.81} & \textbf{76.82\std{1.32}} \\
        & F1 Score & 75.52\std{0.66} & 71.14\std{1.45} & 74.49\std{1.39} & 65.29\std{1.74} 
            & 74.35\std{0.85} & 64.14\std{2.70} & 76.25\std{0.65} & 75.54\std{0.94} & \textbf{76.39\std{1.18}} \\ \midrule

    \multirow{2}{*}{\makecell{\textbf{UCI-HAR } \\ \textit{(6 Classes)}}}

        & Accuracy & 89.74\std{1.08} & 88.44\std{2.02} & 88.86\std{1.65} & 93.08\std{0.95} 
            & 91.44\std{1.01} & 87.78\std{1.10} & 91.65\std{0.74} & 93.72\std{0.73} & \textbf{93.78\std{0.32}} \\
        & F1 Score & 89.70\std{1.10} & 88.34\std{1.98} & 88.80\std{1.67} & 93.19\std{0.88} 
            & 91.47\std{0.98} & 87.72\std{1.10} & 91.61\std{0.75} & 93.69\std{0.76} & \textbf{93.74\std{0.34}} \\

    \bottomrule
    \end{tabular}
    }
\end{table*}

To evaluate the performance of our method on general time series, we follow the design of Medformer~\cite{wang2024medformer} and test it on two human activity recognition (HAR) datasets: FLAAP(13,123 samples, 10 classes)~\cite{kumar2022flaap}  and UCI-HAR(10,299 samples, 6 classes)~\cite{anguita2013public}. Additionally, to conduct a more comprehensive evaluation, following TimeMixer++~\cite{wang2025timemixer++}, we used 10 multivariate datasets from the UEA Time Series Classification Archive (2018) for the assessment of classification tasks.

\begin{wrapfigure}[16]{r}{0.45\textwidth} 
    \centering
    \includegraphics[width=\linewidth]{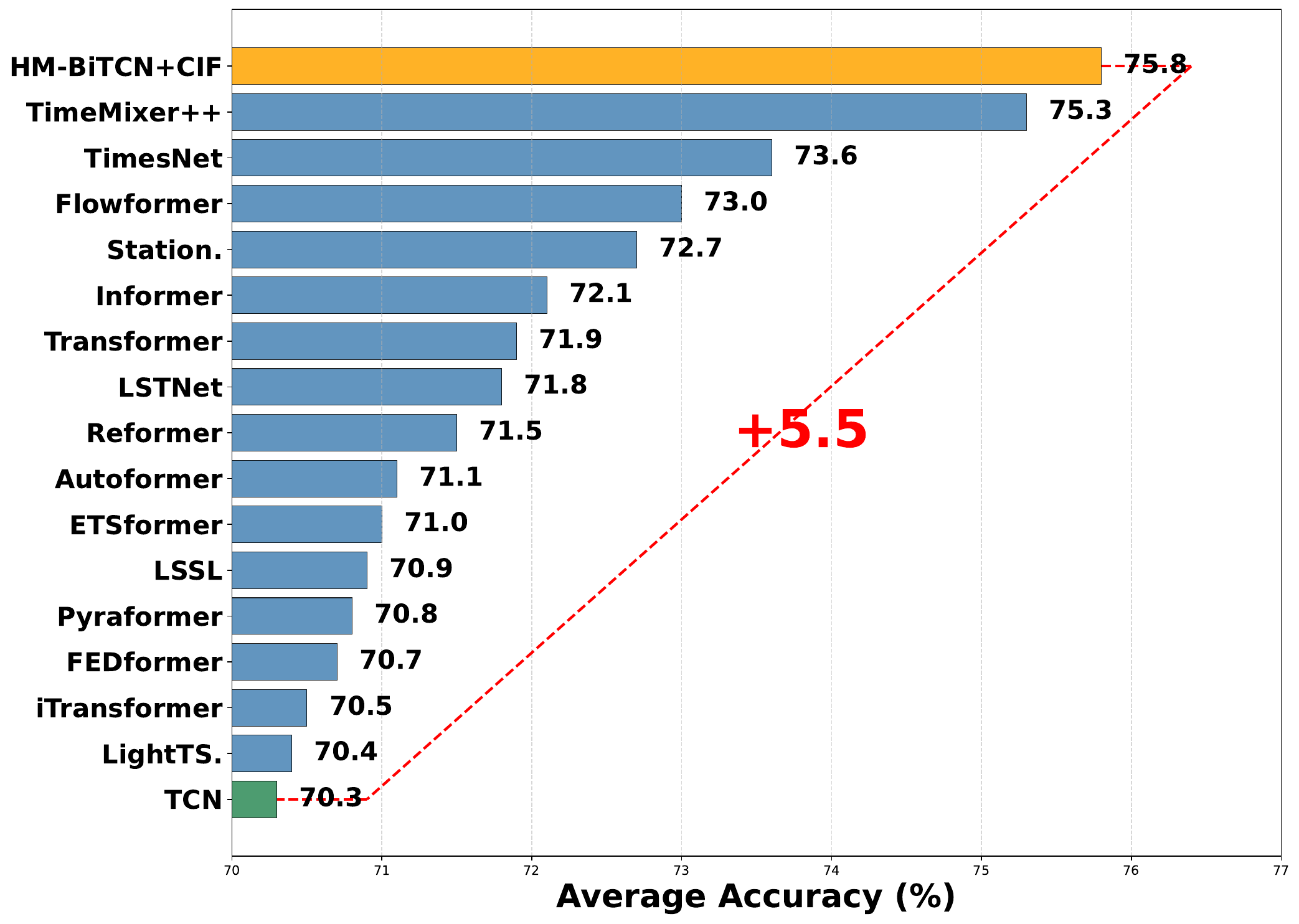}
    \caption{Average accuracy of various methods on the UEA dataset. More details in Appendix~\ref{UEA}.}
    \label{fig:uea_data} 
\end{wrapfigure}
As shown in Table~\ref{tab:human}, and Fig.~\ref{fig:uea_data}, the combination of HM-BiTCN and CIF consistently outperforms other architectures in general time series classification, achieving a 5.5\% improvement over the original TCN and surpassing current SOTA methods. Although CIF was originally designed for MedTS, its integration with HM-BiTCN significantly outperforms Transformer-based models in both medical and general time series classification tasks, demonstrating the effectiveness of our data-centric approach.

The results in Tables~\ref{tab:human}, \ref{tab:UEA}, \ref{tab:abl_differential} and Figure~\ref{fig_APAVA_TDBrain_PTB_APAVA} show that CIF achieves significant improvements in MedTS classification, while the gains on non-medical data are relatively limited. This observation further demonstrates that a data-driven perspective is particularly effective for MedTS classification with physiological characteristics.

\section{Conclusion}
In this work, we propose a simple and effective method for medical time series classification, \textit{Channel-Imposed Fusion} (CIF), which explicitly encodes physiological causal relationships between channels in the feature representations while enhancing the SNR of the original signals. Combined with the simple HM-BiTCN architecture, CIF surpasses existing SOTA methods on multiple medical datasets and performs strongly on general time series classification tasks, demonstrating that data-centric design enables simple models to outperform more complex architectures. More importantly, CIF exemplifies the shift from the traditional \emph{model-centric} paradigm to a \emph{data-centric} perspective, where structured representations grounded in physiological priors are both efficient and scalable for medical time series classification. CIF also exhibits strong transferability and can be seamlessly integrated into mainstream models such as Transformers, enhancing their applicability in medical scenarios. We hope this work encourages the community to reconsider the core of medical time series classification: should it be driven primarily by \emph{data-centric strategies} or by \emph{model-centric design} or both?


\clearpage

\bibliography{iclr2026_conference}
\bibliographystyle{iclr2026_conference}

\clearpage

\appendix
\section*  {Appendix}

\section{Explanation of SNR Optimization via CIF.}
\label{prove_CIF}

Consider the linear combination of the two observed signals:
\begin{equation}
y = a x_1 + b x_2,
\end{equation}
where \(a\) and \(b\) are real coefficients. The observed signals are given by
\begin{equation}
x_1 = s_1 + \epsilon_1, \quad x_2 = s_2 + \epsilon_2,
\end{equation}
with zero-mean signal and noise components:
\[
\mathbb{E}[s_i] = \mathbb{E}[\epsilon_i] = 0, \quad i = 1,2,
\]
and mutually uncorrelated signal and noise components:
\(\mathrm{Cov}(s_i, \epsilon_j) = 0\).

\subsection*{1. Signal Power Calculation}

The power of a zero-mean random signal is given by its variance:
\begin{equation}
P_s = \mathrm{Var}[s] = \mathbb{E}[(s - \mathbb{E}[s])^2] = \mathbb{E}[s^2].
\end{equation}
This is why, for zero-mean signals, the SNR can be expressed as a ratio of variances (or mean-square values) \cite{kay1993fundamentals}.

For the linear combination of signals:
\begin{equation}
\begin{aligned}
\mathrm{Var}(a s_1 + b s_2) &= a^2 \mathrm{Var}(s_1) + b^2 \mathrm{Var}(s_2) + 2ab\,\mathrm{Cov}(s_1,s_2) \\
&= a^2 \sigma_s^2 + b^2 \sigma_s^2 + 2ab (\rho \sigma_s^2) \\
&= \sigma_s^2 \bigl(a^2 + b^2 + 2ab\rho\bigr),
\end{aligned}
\end{equation}
where \(\rho = \mathrm{Corr}(s_1,s_2)\).

\subsection*{2. Noise Power Calculation}

Similarly, the noise power of the linear combination is:
\begin{equation}
\begin{aligned}
\mathrm{Var}(a \epsilon_1 + b \epsilon_2) &= a^2 \mathrm{Var}(\epsilon_1) + b^2 \mathrm{Var}(\epsilon_2) + 2ab\,\mathrm{Cov}(\epsilon_1,\epsilon_2) \\
&= a^2 \sigma_\epsilon^2 + b^2 \sigma_\epsilon^2 + 2ab (\gamma \sigma_\epsilon^2) \\
&= \sigma_\epsilon^2 \bigl(a^2 + b^2 + 2ab\gamma\bigr),
\end{aligned}
\end{equation}
where \(\gamma = \mathrm{Corr}(\epsilon_1,\epsilon_2)\).

\subsection*{3. Output SNR}

Using the definition of SNR as the ratio of signal power to noise power:
\begin{equation}
\mathrm{SNR}_{\text{out}} = \frac{\mathrm{Var}(a s_1 + b s_2)}{\mathrm{Var}(a \epsilon_1 + b \epsilon_2)}
= \frac{\sigma_s^2 (a^2 + b^2 + 2ab\rho)}{\sigma_\epsilon^2 (a^2 + b^2 + 2ab\gamma)}
= \mathrm{SNR}_{\text{in}} \cdot \frac{a^2 + b^2 + 2ab\rho}{a^2 + b^2 + 2ab\gamma},
\end{equation}
where \(\mathrm{SNR}_{\text{in}} = \sigma_s^2 / \sigma_\epsilon^2\).

> \textit{Remark:} The zero-mean property ensures that the variance equals the mean-square value, which is why SNR can be expressed as a ratio of variances~\cite{kay1993fundamentals,haykin2002adaptive}.

\subsection*{4. SNR Improvement Condition}

For SNR improvement relative to individual channels:
\begin{equation}
\frac{a^2 + b^2 + 2ab\rho}{a^2 + b^2 + 2ab\gamma} > 1 \quad \Rightarrow \quad 2ab (\rho - \gamma) > 0.
\label{eq:inequality}
\end{equation}

\begin{itemize}
    \item \textbf{Difference Mode} (\( ab < 0 \)):  
    \(\rho < \gamma\) — suppress correlated noise while possibly attenuating some correlated signal.
    
    \item \textbf{Cooperative Mode} (\( ab > 0 \)):  
    \(\rho > \gamma\) — amplify correlated signals relative to less-correlated noise.
\end{itemize}

\clearpage

\subsection{Evaluation metrics}\label{appendix_evaluation_metrics}
For all methods, the optimizer used is Adam, with a learning rate of 1e-4. The batch size is set to \{32,32,128,128,128\} for the datasets APAVA, TDBrain, ADFD, PTB, and PTB-XL, respectively. Training is conducted for 100 epochs, with early stopping triggered after 10 epochs without improvement in the F1 score on the validation set. We save the model with the best F1 score on the validation set and evaluate it on the test set. We employ six evaluation metrics: accuracy, precision (macro-averaged), recall (macro-averaged), F1 score (macro-averaged), AUROC (macro-averaged), and AUPRC (macro-averaged). Both subject-dependent and subject-independent setups are implemented for different datasets. Each experiment is run with 5 random seeds (41-45) and fixed training, validation, and test sets to compute the average results and standard deviations.

To comprehensively and fairly evaluate the performance of each model in the classification task, we select five evaluation metrics: Accuracy, Precision, Recall, F1 score, and AUROC. The definitions and specific calculation formulas for each metric are presented below:

Accuracy measures the proportion of correct predictions out of the total number of predictions. It's calculated as:
\begin{equation}
    \text { Accuracy }=\frac{\text { Number of correct predictions }}{\text { Total number of predictions }}.
\end{equation}
This metric is useful when the classes are balanced but may be misleading in cases of class imbalance.

Precision focuses on the quality of positive predictions and measures the proportion of correctly predicted positive instances out of all instances predicted as positive. It’s especially useful when false positives need to be minimized. The formula is: 
\begin{equation}
    \text { Precision }=\frac{\text { True Positives }}{\text { True Positives }+ \text { False Positives }}.
\end{equation}

Recall measures the proportion of actual positive instances that were correctly identified. It’s important when false negatives are costly. The formula is:
\begin{equation}
    \text { Recall }=\frac{\text { True Positives }}{\text { True Positives }+ \text { False Negatives }}.
\end{equation}
It shows how well the model captures all relevant instances.

The F1 score is the harmonic mean of precision and recall, balancing the two when one is more important than the other. It’s particularly useful when dealing with imbalanced datasets, as it accounts for both false positives and false negatives. The formula is:
\begin{equation}
    \text { F1 Score }=2 \times \frac{\text { Precision } \times \text { Recall }}{\text { Precision }+ \text { Recall }}.
\end{equation}
It gives a single metric that reflects both precision and recall performance.

The Area Under the Receiver Operating Characteristic Curve (AUROC) measures the
ability of a model to distinguish between classes, defined as
\begin{equation}
    \text{AUROC} = \int_{0}^{1} \text{TPR}(\text{FPR}) \, d(\text{FPR}),
\end{equation}
where
\[
\text{TPR} = \frac{TP}{TP + FN}, \qquad 
\text{FPR} = \frac{FP}{FP + TN}.
\]

The Area Under the Precision--Recall Curve (AUPRC) summarizes the trade-off
between precision and recall across different thresholds, defined as
\begin{equation}
    \text{AUPRC} = \int_{0}^{1} \text{Precision}(\text{Recall}) \, d(\text{Recall}),
\end{equation}
where
\[
\text{Precision} = \frac{TP}{TP + FP}, \qquad 
\text{Recall} = \frac{TP}{TP + FN}.
\]






\clearpage

\section{HM-BiTCN Structure Design and Theoretical Analysis}
\label{app:hmbitcn}
Modeling short-term and long-term dependencies in time series data is challenging. Traditional CNNs excel at capturing local features but have limited receptive fields, hindering long-range dependency learning. Transformer-based methods effectively model long-term dependencies, but their complex design lacks interpretability, which is a key issue in medical time-series classification.To address these limitations, TCNs use causal convolutions for explicit temporal modeling and dilated convolutions to expand the receptive field, overcoming the constraints of traditional CNNs.
Building on the advantages of TCN, we propose the HM-BiTCN, which combines the benefits of dilated convolutions, bidirectional causal convolution, and residual connections. This approach allows for better capture of temporal dependencies while preserving causality. 

\subsection{Dilated Convolution}

Dilated convolution expands the receptive field without significantly increasing computational cost~\cite{yu2015multi}. For a 1D input sequence $ x = [x_1, x_2, \dots, x_T] $, its output is defined as $ y(t) = \sum_{i=0}^{k-1} x(t + i \cdot d) \cdot w(i) $, where $ t $ is the current time step, $ k $ is the kernel size, $ d $ is the dilation factor, and $ w(i) $ is the weight at the $ i $-th position in the kernel. Increasing $ d $ effectively enlarges the receptive field, enabling the network to capture longer-term temporal dependencies. 
When stacking multiple dilated convolutional layers, the receptive field grows progressively. For the $ l $-th layer, the receptive field $ r_l $ can be expressed as $ r_l = k + (k - 1) \sum_{j=1}^{l-1} d_j $, where $ d_j $ is the dilation factor of the $ j $-th layer. By gradually increasing $ d_j $, the network captures temporal dependencies across both global and local scales, offering an effective way to model long-term dependencies in time series.

\subsection{Bidirectional Causal Convolution Structure}

In addition to dilated convolutions, HM-BiTCN introduces a \textit{bidirectional causal convolution structure}, inspired by prior bidirectional temporal modeling approaches~\cite{hanson2018bidirectional, hu2024specslice, yin2025degradation}. Unlike traditional TCNs that use only forward causal convolutions, our architecture applies causal convolutions in both forward and backward directions, enabling the model to capture dependencies from both past and future contexts while strictly preserving causality.
The \textit{forward causal convolution} processes the input sequence $x(t)$ in chronological order, producing output $y_{\text{forward}}(t) = \sum_{i=0}^{k-1} x(t - i \cdot d) \cdot w_{\text{forward}}(i)$, which depends only on current and past inputs. 
For the \textit{backward causal convolution}, we first reverse the input sequence as $x_{\text{flip}}(t) = x(T - t)$, and then apply a causal convolution over this flipped sequence. This ensures that the model captures future-directed dependencies without introducing information leakage. The output is given by $y_{\text{backward}}(t) = \sum_{i=0}^{k-1} x(T - (t - i \cdot d)) \cdot w_{\text{backward}}(i)$.
These two operations are implemented using separate convolutional layers (convforward and convbackward), and their outputs are summed to form the final bidirectional result: $y_{\text{bi}}(t) \;=\; y_{\text{forward}}(t) \;+\; \text{flip}\!\left(y_{\text{backward}}(t)\right)$.
By integrating both directions under strict causality constraints, HM-BiTCN achieves superior temporal dependency modeling compared to unidirectional causal approaches.

\subsection{Multi-Scale Feature Learning and Residual Connections}
To further improve the model's capacity to capture dependencies at different temporal scales, HM-BiTCN incorporates \textit{multi-scale feature learning} and \textit{residual connections}. 
Multi-scale Feature Learning:
In HM-BiTCN, we employ a hierarchy of dilation factors that decrease layer by layer to capture temporal dependencies at multiple scales. Lower layers use larger dilation factors to expand the receptive field, aggregating long‐range information and smoothing short‐term noise in highly redundant medical time series; higher layers use smaller dilation factors to focus on local dependencies and capture fine‐grained features. This coarse‐to‐fine, global‐to‐local design enables the network to extract broad patterns in its initial layers and refine precise details in its later layers, thereby enhancing adaptability across a wide range of time series tasks. Residual connections:
Residual connections~\cite{he2016deep} are introduced between the dilated convolutional layers to facilitate the efficient flow of information through the network. The residual connection is defined as \( y = F(x) + x \), where \( F(x) \) is the convolutional output, and \( x \) is the input. This design alleviates the vanishing gradient problem and improves the overall stability of the network during training.

\section{Pseudocode of CIF Method and Key Components of HM-BiTCN}
\label{app:Pseudocode}

\begin{algorithm}
\caption{CIF Module (Channel-Imposed Fusion)}
\begin{algorithmic}[1]
\Require Input $x_{enc} \in \mathbb{R}^{B \times T \times C}$, hyperparameters $t$, $n$, $a$, $b$
\State $front \gets x_{enc}[:, :, :n]$
\State $back \gets x_{enc}[:, :, -n:]$
\State $x_{new} \gets \text{Clone}(x_{enc})$
\State $added \gets front \cdot a + back \cdot b$
\If{$t > 0$}
    \State $x_{new}[:, :, :n] \gets added$
\Else
    \State $x_{new}[:, :, -n:] \gets added$
\EndIf
\State \Return $x_{new}$
\end{algorithmic}
\end{algorithm}

\begin{algorithm}
\caption{BidirectionalCausalConv}
\begin{algorithmic}[1]
\Require Input $x \in \mathbb{R}^{B \times C \times T}$, kernel size $k$, dilations $d_f$, $d_b$
\State Compute $p_f \gets (k - 1) \cdot d_f$
\State Compute $p_b \gets (k - 1) \cdot d_b$
\State $x_f \gets \text{PadLeft}(x, p_f)$
\State $x_b \gets \text{PadLeft}(\text{Flip}(x), p_b)$
\State $y_f \gets \text{Conv1D}(x_f, \text{dilation}=d_f)$
\State $y_b \gets \text{Flip}(\text{Conv1D}(x_b, \text{dilation}=d_b))$
\State \Return $y_f + y_b$
\end{algorithmic}
\end{algorithm}

\begin{algorithm}
\caption{BidirectionalDilatedConvBlock}
\begin{algorithmic}[1]
\Require Input $x$, channels $C_{in}, C_{out}$, kernel size $k$, dilation $d$
\If{$C_{in} \ne C_{out}$ or final layer}
    \State $res \gets \text{Conv1D}(x, \text{kernel}=1)$
\Else
    \State $res \gets x$
\EndIf
\State $x \gets \text{GELU}(x)$
\State $x \gets \text{BidirectionalCausalConv}(x, k, d, d)$
\State $x \gets \text{GELU}(x)$
\State $x \gets \text{BidirectionalCausalConv}(x, k, d, d)$
\State \Return $x + res$
\end{algorithmic}
\end{algorithm}


\section{Ablation Experiments of the HM-BiTCN Structure}
\label{app:ab_hmbiTCN}

\begin{table}[h]
\centering
\def\arraystretch{1.0}
\caption{The ablation experiments of the HM-BiTCN structure, where “Forward” indicates using only the forward part, and “Backward” indicates using only the backward part.
}
\label{tab:HM-BiTCN-cif}
\resizebox{\textwidth}{!}{%
\begin{tabular}{clc|cc|cccccc}
    \toprule
      \textbf{Datasets} & \textbf{Models} & \textbf{CIF} & \textbf{Forward} & \textbf{Backward} 
      & \multicolumn{1}{c}{\textbf{Accuracy $\uparrow$}} 
      & \multicolumn{1}{c}{\textbf{Precision $\uparrow$}} 
      & \multicolumn{1}{c}{\textbf{Recall $\uparrow$}} 
      & \multicolumn{1}{c}{\textbf{F1 score $\uparrow$}} 
      & \multicolumn{1}{c}{\textbf{AUROC $\uparrow$}} 
      & \multicolumn{1}{c}{\textbf{AUPRC $\uparrow$}} \\
    \midrule
    \multirow{1}{*}{\begin{tabular}[c]{@{}l@{}}\;\makecell{\textbf{APAVA} \\ (2-Classes) } \end{tabular}}
    & \textbf{HM-BiTCN} &  & \checkmark &  & 82.31\std{2.34}& 83.29\std{2.50}& 80.39\std{2.65}& 81.02\std{2.63}& 91.50\std{1.80}& 91.66\std{1.82} \\
    & \textbf{HM-BiTCN} &  &  & \checkmark & 79.45\std{3.51}& 80.69\std{3.04}& 77.14\std{4.47}& 77.58\std{4.75}& 87.95\std{3.82}& 88.41\std{3.74}  \\
    & \textbf{HM-BiTCN} &  & \checkmark & \checkmark  & 82.49\std{1.40} & 82.38\std{1.79} & 81.20\std{1.32} & 81.60\std{1.39} & 91.10\std{1.63} & 91.30\std{1.71} \\
  
\bottomrule

 \toprule
  \multirow{1}{*}{\begin{tabular}[c]{@{}l@{}}\;\makecell{\textbf{APAVA} \\ (2-Classes) } \end{tabular}}
    & \textbf{HM-BiTCN} & \checkmark & \checkmark &  & 80.43\std{5.60}& 80.46\std{5.23}& 79.56\std{5.98}& 79.50\std{5.97}& 89.23\std{4.44}& 89.62\std{4.25}  \\
    & \textbf{HM-BiTCN} & \checkmark &  & \checkmark & 79.39\std{3.44}& 79.49\std{3.76}& 78.09\std{2.94}& 78.35\std{3.33}& 87.62\std{3.09}& 88.11\std{2.91}  \\
    & \textbf{HM-BiTCN} & \checkmark & \checkmark & \checkmark  &  85.16\std{1.55} &  84.76\std{1.62} &  85.33\std{1.27} &  84.82\std{1.49} &  94.06\std{1.07} &  94.21\std{0.99} \\
\bottomrule

 \toprule
  
    \midrule
    \multirow{1}{*}{\begin{tabular}[c]{@{}l@{}}\;\makecell{\textbf{ADFTD} \\ (3-Classes) } \end{tabular}}
    & \textbf{HM-BiTCN} &  & \checkmark &  & 53.32\std{1.35}& 52.01\std{1.54}& 51.46\std{2.19}& 51.21\std{1.99}& 70.78\std{1.78}& 53.16\std{2.20} \\
    & \textbf{HM-BiTCN} & &  & \checkmark & 52.80\std{1.18}& 50.16\std{0.77}& 49.23\std{1.22}& 49.24\std{1.02}& 68.65\std{0.71}& 49.95\std{0.97} \\
    & \textbf{HM-BiTCN} &  & \checkmark & \checkmark  & 52.05\std{2.22} & 50.45\std{3.00} & 50.40\std{2.55} & 49.48\std{2.70} & 69.43\std{2.84} & 50.99\std{3.15} \\
  

  
    \midrule
    \multirow{1}{*}{\begin{tabular}[c]{@{}l@{}}\;\makecell{\textbf{ADFTD} \\ (3-Classes) } \end{tabular}}
    & \textbf{HM-BiTCN} & \checkmark & \checkmark &  & 56.06\std{0.47}& 53.21\std{1.03}& 53.54\std{1.36}& 52.82\std{1.33}& 72.93\std{0.88}& 55.71\std{1.03} \\
    & \textbf{HM-BiTCN} & \checkmark &  & \checkmark & 56.54\std{1.33}& 54.28\std{0.96}& 54.63\std{1.06}& 53.91\std{1.11}& 73.46\std{1.17}& 56.12\std{1.61} \\
    & \textbf{HM-BiTCN} & \checkmark & \checkmark & \checkmark    &58.56\std{0.93} & 55.65\std{0.81} &  55.86\std{0.79} &  55.42\std{0.82} &  76.07\std{0.59} &  59.75\std{0.67} \\
  
\bottomrule

\toprule
  
    \midrule
    \multirow{1}{*}{\begin{tabular}[c]{@{}l@{}}\;\makecell{\textbf{TDBrain} \\ (2-Classes) } \end{tabular}}
    & \textbf{HM-BiTCN} &  & \checkmark &  & 87.23\std{2.87}& 87.75\std{2.48}& 87.23\std{2.87}& 87.17\std{2.93}& 95.55\std{1.69}& 95.73\std{1.60} \\
    & \textbf{HM-BiTCN} & &  & \checkmark & 86.92\std{3.46}& 87.41\std{3.17}& 86.92\std{3.46}& 86.86\std{3.51}& 95.28\std{1.78}& 95.42\std{1.70}  \\
    & \textbf{HM-BiTCN} &  & \checkmark & \checkmark  & 84.90\std{2.60} & 86.02\std{2.00} & 84.90\std{2.60} & 84.76\std{2.74} & 93.94\std{1.92} & 94.20\std{1.85} \\
  

  
    \midrule
    \multirow{1}{*}{\begin{tabular}[c]{@{}l@{}}\;\makecell{\textbf{TDBrain} \\ (2-Classes) } \end{tabular}}
    & \textbf{HM-BiTCN} & \checkmark & \checkmark &  & 93.29\std{1.73}& 93.34\std{1.73}& 93.29\std{1.73}& 93.29\std{1.73}& 98.50\std{0.63}& 98.56\std{0.60} \\
    & \textbf{HM-BiTCN} & \checkmark &  & \checkmark  & 93.69\std{1.52}& 93.83\std{1.42}& 93.69\std{1.52}& 93.68\std{1.53}& 98.56\std{0.67}& 98.59\std{0.64}   \\
    & \textbf{HM-BiTCN} & \checkmark & \checkmark & \checkmark  &  93.13\std{1.41} &  93.33\std{1.37} &  93.13\std{1.41} &  93.12\std{1.42} &  98.62\std{0.66} &  98.68\std{0.63} \\
  
\bottomrule

\toprule
  
    \midrule
    \multirow{1}{*}{\begin{tabular}[c]{@{}l@{}}\;\makecell{\textbf{PTB} \\ (2-Classes) } \end{tabular}}
    & \textbf{HM-BiTCN} &  & \checkmark &  & 82.56\std{1.74}& 86.16\std{1.51}& 74.91\std{2.88}& 77.24\std{2.92}& 95.69\std{0.64}& 94.56\std{0.76}  \\
    & \textbf{HM-BiTCN} &  &  & \checkmark & 81.07\std{4.24}& 85.36\std{2.71}& 72.50\std{6.59}& 74.33\std{6.71}& 92.83\std{2.38}& 91.28\std{2.79}  \\
    & \textbf{HM-BiTCN} &  & \checkmark & \checkmark  & 81.87\std{1.87} & 86.50\std{1.24} & 73.49\std{2.90} & 75.84\std{3.20} & 94.20\std{0.29} & 93.04\std{0.45} \\
  

  
    \midrule
    \multirow{1}{*}{\begin{tabular}[c]{@{}l@{}}\;\makecell{\textbf{PTB} \\ (2-Classes) } \end{tabular}}
    & \textbf{HM-BiTCN} & \checkmark & \checkmark &  & 87.33\std{1.41}& 90.26\std{1.24}& 81.64\std{2.04}& 84.19\std{1.97}& 96.21\std{1.30}& 95.67\std{1.52}  \\
    & \textbf{HM-BiTCN} & \checkmark &  & \checkmark & 84.35\std{2.28}& 87.42\std{2.07}& 77.54\std{3.30}& 79.98\std{3.41}& 91.25\std{1.92}& 90.42\std{2.28}  \\
    & \textbf{HM-BiTCN} & \checkmark & \checkmark & \checkmark   &  88.29\std{1.45} &  90.66\std{1.48} &  83.21\std{2.02} &  85.59\std{1.96} &  94.28\std{0.93} &  93.78\std{1.11} \\
  
\bottomrule

\toprule
  
    \midrule
    \multirow{1}{*}{\begin{tabular}[c]{@{}l@{}}\;\makecell{\textbf{FLAAP} \\ (10-Classes) } \end{tabular}}
    & \textbf{HM-BiTCN} &  & \checkmark &  & 70.81\std{2.31}& 72.58\std{1.33}& 69.81\std{2.79}& 70.07\std{2.24}& 95.89\std{0.28}& 76.90\std{1.21} \\
    & \textbf{HM-BiTCN} &  &  & \checkmark & 70.29\std{2.04}& 72.77\std{2.09}& 68.86\std{2.39}& 69.56\std{1.98}& 95.61\std{0.26}& 76.56\std{1.56}  \\
    & \textbf{HM-BiTCN} &  & \checkmark & \checkmark  & 76.08\std{0.81}& 76.05\std{0.83}& 75.95\std{0.84}& 75.54\std{0.94}& 96.49\std{0.10}& 81.19\std{0.65} \\
  

  
    \midrule
    \multirow{1}{*}{\begin{tabular}[c]{@{}l@{}}\;\makecell{\textbf{FLAAP} \\ (10-Classes) } \end{tabular}}
    & \textbf{HM-BiTCN} & \checkmark & \checkmark &  & 72.30\std{1.50}& 72.98\std{1.61}& 71.65\std{1.35}& 71.54\std{1.50}& 95.92\std{0.55}& 77.87\std{2.27}  \\
    & \textbf{HM-BiTCN} & \checkmark &  & \checkmark & 72.81\std{1.04}& 74.05\std{0.80}& 71.86\std{1.25}& 72.12\std{1.17}& 96.20\std{0.21}& 79.22\std{1.25}  \\
    & \textbf{HM-BiTCN} & \checkmark & \checkmark & \checkmark   & 76.82\std{1.32}& 77.38\std{0.85}& 76.52\std{1.24}& 76.39\std{1.18}& 96.48\std{0.06}& 81.77\std{0.81} \\
  
\bottomrule

\toprule
    \midrule
    \multirow{1}{*}{\begin{tabular}[c]{@{}l@{}}\;\makecell{\textbf{UCI-HAR} \\ (6-Classes) } \end{tabular}}
    & \textbf{HM-BiTCN} &  & \checkmark &  & 91.94\std{0.98}& 92.36\std{0.90}& 92.02\std{0.96}& 91.98\std{0.93}& 99.30\std{0.08}& 97.31\std{0.47} \\
    & \textbf{HM-BiTCN} &  &  & \checkmark 
    & 93.03\std{0.62}& 93.28\std{0.63}& 93.12\std{0.60}& 93.05\std{0.62}& 99.36\std{0.19}& 97.72\std{0.46}\\
    & \textbf{HM-BiTCN} &  & \checkmark & \checkmark  & 93.72\std{0.73}& 94.02\std{0.72}& 93.75\std{0.70}& 93.69\std{0.76}& 99.60\std{0.09}& 98.31\std{0.40} \\
  
    \midrule
    \multirow{1}{*}{\begin{tabular}[c]{@{}l@{}}\;\makecell{\textbf{UCI-HAR} \\ (6-Classes) } \end{tabular}}
    & \textbf{HM-BiTCN} & \checkmark & \checkmark &  & 92.18\std{0.45}& 92.42\std{0.47}& 92.21\std{0.44}& 92.14\std{0.44}& 99.17\std{0.11}& 97.04\std{0.15} \\
    & \textbf{HM-BiTCN} & \checkmark &  & \checkmark & 92.62\std{0.90}& 92.88\std{0.86}& 92.68\std{0.88}& 92.63\std{0.89}& 99.25\std{0.18}& 97.15\std{0.57}  \\
    & \textbf{HM-BiTCN} & \checkmark & \checkmark & \checkmark   & 93.78\std{0.32}& 94.08\std{0.26}& 93.79\std{0.32}& 93.74\std{0.34}& 99.34\std{0.19}& 97.60\std{0.46} \\
  
\bottomrule

\end{tabular}
}
\end{table}

From the table \ref{tab:HM-BiTCN-cif}, it can be observed that when both the Forward and Backward parts of the HM-BiTCN structure are used simultaneously, the performance drops significantly compared to using only one of them individually. We speculate that this is mainly due to the presence of substantial noise within medical time-series data. When both parts of the structure are applied at the same time, it is akin to capturing noise from two different directions simultaneously. Instead of enhancing the representation, this leads to noise accumulation, which ultimately results in degraded performance.

However, after processing the data with CIF to improve the signal-to-noise ratio, the combination of the Forward and Backward parts of the HM-BiTCN structure eventually outperforms the use of either part alone. This result strongly demonstrates the feature-capturing capability of the HM-BiTCN when both directions are utilized together. It indicates that once noise interference is effectively reduced, the bidirectional structure of HM-BiTCN can better leverage its strengths, thereby improving overall performance.

Further observations show that using only the Forward part of HM-BiTCN outperforms the Backward part. This is closely related to the inherent unidirectionality of medical time-series signals such as EEG and ECG, where information typically propagates forward in time (e.g., neural signal transmission in EEG or atrial-to-ventricular activation in ECG). Such characteristics enable the Forward structure to capture key features and temporal evolution more effectively, yielding better performance. This finding not only deepens the understanding of medical signal processing but also provides insights for optimizing HM-BiTCN in related applications.

To evaluate the performance of our method on general time series, we follow the design of Medformer~\cite{wang2024medformer} and test it on two human activity recognition (HAR) datasets: FLAAP(13,123 samples, 10 classes)~\cite{kumar2022flaap}  and UCI-HAR(10,299 samples, 6 classes)~\cite{anguita2013public}.

Additionally, on the non-medical datasets FLAAP and UCI-HAR, we observed that integrating the bidirectional structure significantly improves performance. This indicates that in high-SNR scenarios, bidirectional modeling can more effectively capture both forward and backward feature information, enhancing overall model performance. In contrast, CIF provides relatively limited gains on these high-SNR datasets. This observation further highlights the design advantage of CIF: it is specifically tailored for low-SNR medical time series, explicitly fusing inter-channel physiological information to enhance signal quality and discriminative power, while its marginal benefit is smaller for low-noise non-medical data. Overall, these findings not only reveal the differential adaptability of model architectures under varying data characteristics but also underscore the unique value of CIF in complex medical scenarios.


\section{Further exploration of physiological structures}
\label{app:PSF}


\begin{wrapfigure}[20]{r}{0.5\textwidth} 
    \centering
    \includegraphics[width=\linewidth]{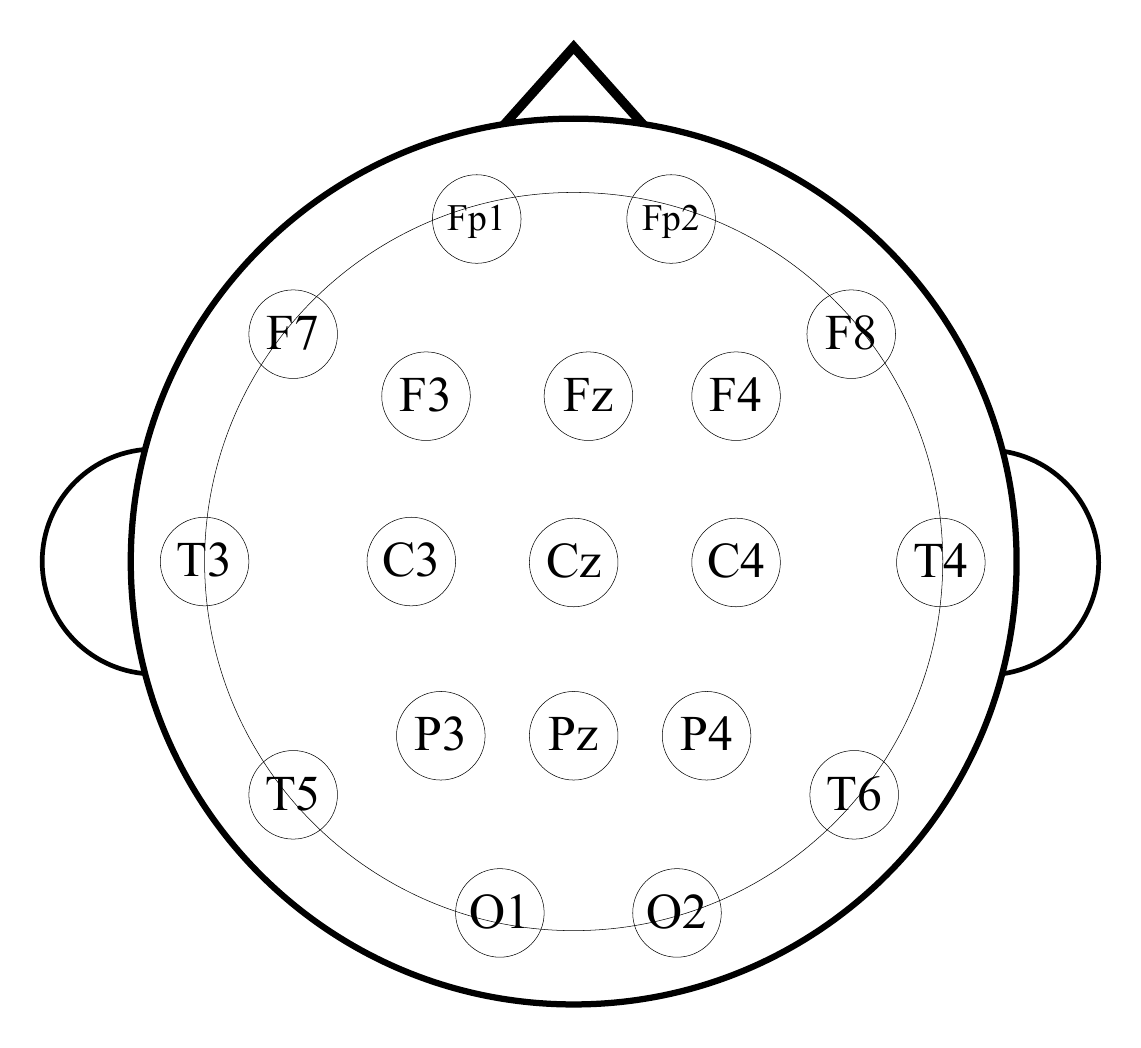}
    \caption{Physiological Placement Diagram of EEG Channels.}
    \label{fig_PSF}
\end{wrapfigure}

The parameters $(a, b, t, n)$ in CIF are explicit hyperparameters that can be directly set and adjusted based on experience. For example, Figure.~\ref{fig_PSF} shows the corresponding locations of EEG channels on the human brain, we have adjusted the AFAVA dataset, which comprises 16 channels: \texttt{C3}, \texttt{C4}, \texttt{F3}, \texttt{F4}, \texttt{F7}, \texttt{F8}, \texttt{Fp1}, \texttt{Fp2}, \texttt{O1}, \texttt{O2}, \texttt{P3}, \texttt{P4}, \texttt{T3}, \texttt{T4}, \texttt{T5}, and \texttt{T6}. For the first six channels, we performed pairwise fusion as follows:
\begin{align*}
C3_{\text{new}} &= a \cdot C3 + b \cdot C4, \\
F3_{\text{new}} &= a \cdot F3 + b \cdot F4, \\
F7_{\text{new}} &= a \cdot F7 + b \cdot F8.
\end{align*}
Here, the channels \texttt{C3}, \texttt{F3}, and \texttt{F7} exhibit physiological symmetry with \texttt{C4}, \texttt{F4}, and \texttt{F8}, respectively. We denote this type of fusion as Physiological Symmetry Fusion (PSF). In contrast, the channel fusion previously described in our paper, which was adjusted via the hyperparameter \( n \), lacked physiological symmetry and is referred to as Random Fusion (RF).

\begin{table}[htbp]
\centering
\def\arraystretch{1.0}
\caption{\textbf{Results of Subject-Independent Setup. APAVA Dataset} 
}
\label{tab:APAVA_2}
\resizebox{\textwidth}{!}{%
\begin{tabular}{clcccccc}
    \toprule
   
      \textbf{Datasets} & \textbf{Models} & \multicolumn{1}{c}{\textbf{Accuracy $\uparrow$}} & \multicolumn{1}{c}{\textbf{Precision $\uparrow$}} & \multicolumn{1}{c}{\textbf{Recall $\uparrow$}} & \multicolumn{1}{c}{\textbf{F1 score $\uparrow$}} & \multicolumn{1}{c}{\textbf{AUROC $\uparrow$}} & \multicolumn{1}{c}{\textbf{AUPRC $\uparrow$}} \\
    \midrule
    \multirow{2}{*}{\begin{tabular}[c]{@{}l@{}}\;\makecell{\textbf{APAVA} \\ (2-Classes) } \end{tabular}} 
  
    & HM-BiTCN + CIF(RF) & 85.16\std{1.55} & 84.76\std{1.62} & 85.33\std{1.27} & 84.82\std{1.49} & 94.06\std{1.07} & 94.21\std{0.99} \\

      & HM-BiTCN + CIF(PSF) & \textbf{86.23\std{2.09}}& \textbf{85.82\std{2.14}}& \textbf{86.04\std{2.06}}& \textbf{85.83\std{2.12}}& \textbf{94.59\std{1.08}}& \textbf{94.64\std{1.08}}  \\
    \midrule
    \bottomrule
\end{tabular}
}
\end{table}

The results in the table \ref{tab:APAVA_2} reveal that explicit fusion leveraging the prior knowledge of channels can more effectively integrate channel features, thereby yielding more accurate classification outcomes. Many previous methods, especially various general time series models, are unable to incorporate such medical prior knowledge in a "controllable" manner.



\section{Results on General Time Series}
\label{UEA}

\begin{table}[htbp]
  \caption{Full results for the classification task. $\ast.$ in the Transformers indicates the name of $\ast$former. We report the classification accuracy (\%) as the result.  }\label{tab:UEA}
  \vskip 0.05in
  \centering
  \resizebox{\columnwidth}{!}{
  \begin{threeparttable}
  \begin{small}
  \renewcommand{\multirowsetup}{\centering}
  \setlength{\tabcolsep}{0.1pt}
  \begin{tabular}{c|ccccccccccccccccccccccccccccccccccc}
    \toprule
    \multirow{3}{*}{\scalebox{0.8}{Datasets / Models}}  & \multicolumn{3}{c}{\scalebox{0.8}{RNN}}& \scalebox{0.8}{TCN} & \multicolumn{10}{c}{\scalebox{0.8}{Transformers}} & \multicolumn{3}{c}{\scalebox{0.8}{MLP}}  & \multicolumn{1}{c}{\scalebox{0.8}{CNN}}\\
    \cmidrule(lr){2-4}\cmidrule(lr){5-5}\cmidrule(lr){6-15}\cmidrule(lr){16-18}\cmidrule(lr){19-19}
    & \scalebox{0.6}{LSTM} & \scalebox{0.6}{LSTNet} & \scalebox{0.6}{LSSL} & \scalebox{0.6}{TCN} & \scalebox{0.8}{Trans.} & \scalebox{0.8}{Re.} & \scalebox{0.8}{In.} & \scalebox{0.8}{Pyra.} & \scalebox{0.8}{Auto.} & \scalebox{0.8}{Station.} &  \scalebox{0.8}{FED.} & \scalebox{0.8}{ETS.} & \scalebox{0.8}{Flow.} &  \scalebox{0.8}{iTrans.}& \scalebox{0.7}{DLinear} & \scalebox{0.7}{LightTS.}&  \scalebox{0.8}{TiDE} &  \scalebox{0.8}{TimesNet}  &  \scalebox{0.8}{
  \parbox{2cm}{\centering Time\\Mixer++}} & \scalebox{0.8}{\parbox{2cm}{\centering\textbf{HM-BiTCN}}}  & \scalebox{0.8}{\parbox{2cm}{\centering\textbf{HM-BiTCN}\\\textbf{+CIF}}} \\
	& \scalebox{0.7}{\citeyearpar{hochreiter1997long}} & 
	\scalebox{0.7}{\citeyearpar{2018Modeling}} & 
	\scalebox{0.7}{\citeyearpar{gu2022efficiently}} & 
	\scalebox{0.7}{\citeyearpar{franceschi2019unsupervised}} & \scalebox{0.7}{\citeyearpar{vaswani2017attention}} & 
	\scalebox{0.7}{\citeyearpar{kitaev2019reformer}} & \scalebox{0.7}{\citeyearpar{zhou2021informer}} & \scalebox{0.7}{\citeyearpar{liu2021pyraformer}} &
	\scalebox{0.7}{\citeyearpar{wu2021autoformer}} & 
	\scalebox{0.7}{\citeyearpar{liu2022non}} &
	\scalebox{0.7}{\citeyearpar{zhou2022fedformer}} & \scalebox{0.7}{\citeyearpar{woo2022etsformer}} & \scalebox{0.7}{\citeyearpar{wu2022flowformer}}& \scalebox{0.7}{\citeyearpar{liu2023itransformer}} & 
	\scalebox{0.7}{\citeyearpar{dlinear}} & \scalebox{0.7}{\citeyearpar{lightts}} & \scalebox{0.7}{\citeyearpar{das2023long}} & \scalebox{0.7}{\citeyearpar{wu2022timesnet}}& \scalebox{0.7}{\citeyearpar{wang2025timemixer++}} & \scalebox{0.7}{\textbf{(Ours)}} & \scalebox{0.7}{\textbf{(Ours)}} \\
    \toprule
	\scalebox{0.7}{EthanolConcentration}  & \scalebox{0.8}{32.3} & \scalebox{0.8}{39.9} & \scalebox{0.8}{31.1}&  \scalebox{0.8}{28.9} & \scalebox{0.8}{32.7} &\scalebox{0.8}{31.9} &\scalebox{0.8}{31.6}   &\scalebox{0.8}{30.8} &\scalebox{0.8}{31.6} &\scalebox{0.8}{32.7} & \scalebox{0.8}{28.1}&\scalebox{0.8}{31.2}  & \scalebox{0.8}{33.8} & \scalebox{0.8}{28.1}& \scalebox{0.8}{32.6} &\scalebox{0.8}{29.7} & \scalebox{0.8}{27.1}& \scalebox{0.8}{35.7} & \scalebox{0.8}{39.9} & \scalebox{0.8}{31.9}  & \scalebox{0.8}{32.3}\\
	\scalebox{0.7}{FaceDetection}& \scalebox{0.8}{57.7} & \scalebox{0.8}{65.7} & \scalebox{0.8}{66.7} & \scalebox{0.8}{52.8} & \scalebox{0.8}{67.3} & \scalebox{0.8}{68.6} &\scalebox{0.8}{67.0} &\scalebox{0.8}{65.7} &\scalebox{0.8}{68.4} &\scalebox{0.8}{68.0} &\scalebox{0.8}{66.0} & \scalebox{0.8}{66.3} & \scalebox{0.8}{67.6} & \scalebox{0.8}{66.3}&\scalebox{0.8}{68.0} &\scalebox{0.8}{67.5} & \scalebox{0.8}{65.3}& \scalebox{0.8}{68.6} & \scalebox{0.8}{71.8} & \scalebox{0.8}{66.8}  & \scalebox{0.8}{67.2} \\
	\scalebox{0.7}{Handwriting} & \scalebox{0.8}{15.2} & \scalebox{0.8}{25.8} & \scalebox{0.8}{24.6} & \scalebox{0.8}{53.3} & \scalebox{0.8}{32.0} & \scalebox{0.8}{27.4} &\scalebox{0.8}{32.8} &\scalebox{0.8}{29.4} &\scalebox{0.8}{36.7} &\scalebox{0.8}{31.6} &\scalebox{0.8}{28.0} &  \scalebox{0.8}{32.5} & \scalebox{0.8}{33.8} & \scalebox{0.8}{24.2}& \scalebox{0.8}{27.0} &\scalebox{0.8}{26.1} & \scalebox{0.8}{23.2}& \scalebox{0.8}{32.1} & \scalebox{0.8}{26.5}
    & \scalebox{0.8}{49.5}& \scalebox{0.8}{51.2} \\
	\scalebox{0.7}{Heartbeat}  & \scalebox{0.8}{72.2} & \scalebox{0.8}{77.1} & \scalebox{0.8}{72.7}& \scalebox{0.8}{75.6} & \scalebox{0.8}{76.1} & \scalebox{0.8}{77.1} &\scalebox{0.8}{80.5} &\scalebox{0.8}{75.6} &\scalebox{0.8}{74.6} &\scalebox{0.8}{73.7} &\scalebox{0.8}{73.7} &  \scalebox{0.8}{71.2} & \scalebox{0.8}{77.6} & \scalebox{0.8}{75.6}& \scalebox{0.8}{75.1} &\scalebox{0.8}{75.1} & \scalebox{0.8}{74.6}& \scalebox{0.8}{78.0}  & \scalebox{0.8}{79.1} & \scalebox{0.8}{74.6}   & \scalebox{0.8}{77.5} \\
	\scalebox{0.7}{JapaneseVowels}  & \scalebox{0.8}{79.7} & \scalebox{0.8}{98.1} & \scalebox{0.8}{98.4} & \scalebox{0.8}{98.9} & \scalebox{0.8}{98.7} & \scalebox{0.8}{97.8} &\scalebox{0.8}{98.9} &\scalebox{0.8}{98.4} &\scalebox{0.8}{96.2} &\scalebox{0.8}{99.2} &\scalebox{0.8}{98.4} & \scalebox{0.8}{95.9} &  \scalebox{0.8}{98.9} & \scalebox{0.8}{96.6}& \scalebox{0.8}{96.2} &\scalebox{0.8}{96.2} & \scalebox{0.8}{95.6}& \scalebox{0.8}{98.4} & \scalebox{0.8}{97.9} & \scalebox{0.8}{97.8} & \scalebox{0.8}{98.3} \\
	\scalebox{0.7}{PEMS-SF}& \scalebox{0.8}{39.9} & \scalebox{0.8}{86.7} & \scalebox{0.8}{86.1}& \scalebox{0.8}{68.8} & \scalebox{0.8}{82.1} & \scalebox{0.8}{82.7} &\scalebox{0.8}{81.5} &\scalebox{0.8}{83.2} &\scalebox{0.8}{82.7} &\scalebox{0.8}{87.3} &\scalebox{0.8}{80.9} & \scalebox{0.8}{86.0} &  \scalebox{0.8}{83.8} & \scalebox{0.8}{87.9}& \scalebox{0.8}{75.1} &\scalebox{0.8}{88.4} & \scalebox{0.8}{86.9}& \scalebox{0.8}{89.6} & \scalebox{0.8}{91.0} & \scalebox{0.8}{82.6} & \scalebox{0.8}{86.1}\\
	\scalebox{0.7}{SelfRegulationSCP1} & \scalebox{0.8}{68.9} & \scalebox{0.8}{84.0} & \scalebox{0.8}{90.8} & \scalebox{0.8}{84.6} & \scalebox{0.8}{92.2} & \scalebox{0.8}{90.4} &\scalebox{0.8}{90.1} &\scalebox{0.8}{88.1} &\scalebox{0.8}{84.0} &\scalebox{0.8}{89.4} &\scalebox{0.8}{88.7} & \scalebox{0.8}{89.6} & \scalebox{0.8}{92.5} & \scalebox{0.8}{90.2}& \scalebox{0.8}{87.3} &\scalebox{0.8}{89.8} & \scalebox{0.8}{89.2}& \scalebox{0.8}{91.8}  & \scalebox{0.8}{93.1} & \scalebox{0.8}{89.7}& \scalebox{0.8}{91.1} \\
    \scalebox{0.7}{SelfRegulationSCP2}& \scalebox{0.8}{46.6} & \scalebox{0.8}{52.8} & \scalebox{0.8}{52.2} & \scalebox{0.8}{55.6} & \scalebox{0.8}{53.9} & \scalebox{0.8}{56.7} &\scalebox{0.8}{53.3} &\scalebox{0.8}{53.3} &\scalebox{0.8}{50.6} &\scalebox{0.8}{57.2} &\scalebox{0.8}{54.4} & \scalebox{0.8}{55.0} &  \scalebox{0.8}{56.1} & \scalebox{0.8}{54.4}& \scalebox{0.8}{50.5} &\scalebox{0.8}{51.1} & \scalebox{0.8}{53.4}& \scalebox{0.8}{57.2} & \scalebox{0.8}{65.6} & \scalebox{0.8}{61.6} & \scalebox{0.8}{62.2}\\
    \scalebox{0.7}{SpokenArabicDigits} & \scalebox{0.8}{31.9} & \scalebox{0.8}{100.0} & \scalebox{0.8}{100.0} & \scalebox{0.8}{95.6} & \scalebox{0.8}{98.4} & \scalebox{0.8}{97.0} &\scalebox{0.8}{100.0} &\scalebox{0.8}{99.6} &\scalebox{0.8}{100.0} &\scalebox{0.8}{100.0} &\scalebox{0.8}{100.0} & \scalebox{0.8}{100.0} &  \scalebox{0.8}{98.8} & \scalebox{0.8}{96.0}& \scalebox{0.8}{81.4} &\scalebox{0.8}{100.0} & \scalebox{0.8}{95.0}& \scalebox{0.8}{99.0} & \scalebox{0.8}{99.8} & \scalebox{0.8}{99.5}& \scalebox{0.8}{99.6}\\
    \scalebox{0.7}{UWaveGestureLibrary} & \scalebox{0.8}{41.2} & \scalebox{0.8}{87.8} & \scalebox{0.8}{85.9} & \scalebox{0.8}{88.4} & \scalebox{0.8}{85.6} & \scalebox{0.8}{85.6} &\scalebox{0.8}{85.6} &\scalebox{0.8}{83.4} &\scalebox{0.8}{85.9} &\scalebox{0.8}{87.5} &\scalebox{0.8}{85.3} & \scalebox{0.8}{85.0} &  \scalebox{0.8}{86.6} & \scalebox{0.8}{85.9}& \scalebox{0.8}{82.1} &\scalebox{0.8}{80.3} & \scalebox{0.8}{84.9}& \scalebox{0.8}{85.3} & \scalebox{0.8}{88.2}  & \scalebox{0.8}{92.1}& \scalebox{0.8}{92.8} \\
    \midrule
    \scalebox{0.8}{Average Accuracy}  & \scalebox{0.8}{48.6} & \scalebox{0.8}{71.8} & \scalebox{0.8}{70.9} & \scalebox{0.8}{70.3} & \scalebox{0.8}{71.9} & \scalebox{0.8}{71.5} &\scalebox{0.8}{72.1} &\scalebox{0.8}{70.8} &\scalebox{0.8}{71.1} &\scalebox{0.8}{72.7} &\scalebox{0.8}{70.7} & \scalebox{0.8}{71.0} &{\scalebox{0.8}{73.0}} & \scalebox{0.8}{70.5}& \scalebox{0.8}{67.5} &\scalebox{0.8}{70.4} & \scalebox{0.8}{69.5}& \scalebox{0.8}{73.6} & \scalebox{0.8}{75.3} & \scalebox{0.8}{74.6} & \scalebox{0.8}{\textbf{75.8}} \\
	\bottomrule
  \end{tabular}
    \end{small}
  \end{threeparttable}
  }
\end{table}

We compared our method with various approaches on the general time series UEA dataset. The results of these methods were provided by TimeMixer++~\cite{wang2025timemixer++}. Experimental results show that CIF can enhance the performance of HM-BiTCN on general time series classification tasks. Moreover, the combination of HM-BiTCN with CIF achieves SOTA performance.

\newpage

\clearpage
\section{Limitations and Future Work}
\label{limitations}

\textbf{Limitations:}  

Biomedical time series exhibit complex modal characteristics, which lead to significant efficiency bottlenecks when manually adjusting the prior parameters (t, n, a, b) with clear medical interpretations in the CIF model based on empirical experience. This limitation highlights the urgent need for developing novel automated hyperparameter optimization frameworks.

\textbf{Future Work:}  

We plan to explore a more universal and generalizable time-series analysis approach, incorporating domain knowledge, structural modeling, and automated hyperparameter optimization. This integration should foster both deeper theoretical insights and stronger practical applicability, providing robust solutions for real-world medical problems.

Furthermore, incorporating domain-specific prior knowledge into medical time-series analysis can more precisely reveal and model relationships between channels. By integrating medical expertise, clinical experience, and existing pathological data, the interpretability and predictive performance of models can be enhanced, thereby supporting clinical decision-making and interventions. On this basis, frequency-domain analysis~\cite{hu2025beta, nason1999wavelets, yi2025survey} offers an additional perspective: by applying Fourier transform or wavelet decomposition to the signals, physiological features at different frequency components can be identified, revealing patterns that are difficult to capture in the time domain. This is particularly valuable for noise reduction, extraction of periodic signals, and detection of pathological events, and can also provide richer feature representations for model inputs. Future research could further explore how to combine time-domain and frequency-domain information, integrating domain priors to improve the accuracy and robustness of intelligent medical analytics.

Finally, we must acknowledge that the development trends in the field of artificial intelligence highlight the importance of architectural innovation. Future research should focus on designing novel architectures that align more closely with the CIF method, combining the strengths of existing models. For example, the local feature extraction capabilities of CNNs~\cite{lecun1989backpropagation}, the temporal stability of TCNs~\cite{bai2018empirical} for long sequences, the long-term dependency modeling of RNNs~\cite{rumelhart1986learning} and LSTMs~\cite{hochreiter1997long}, the global modeling efficiency of Transformers~\cite{vaswani2017attention}, the resource-efficient computation of Mamba~\cite{gu2023mamba}, and the hybrid recurrence-attention structure of RWKV~\cite{peng2023rwkv}. By adapting and integrating these methods, we aim to build a powerful model that is not only deeply compatible with the CIF framework but also capable of efficiently handling complex medical time-series data.


\end{document}